\documentclass[10pt,twocolumn,letterpaper]{article}

\usepackage[pagenumbers]{cvpr} 

\usepackage{graphicx}
\usepackage{amsmath}
\usepackage{amssymb}
\usepackage{booktabs}
\usepackage{epigraph}
\usepackage{diagbox}
\usepackage{xfrac}  
\usepackage[accsupp]{axessibility} 

\usepackage{array,multirow,makecell,tabularx}

\usepackage{float}
\usepackage{stfloats}

\usepackage[pagebackref,breaklinks,colorlinks]{hyperref}

\usepackage[capitalize]{cleveref}
\crefname{section}{Sec.}{Secs.}
\Crefname{section}{Section}{Sections}
\Crefname{table}{Table}{Tables}
\crefname{table}{Tab.}{Tabs.}
\Crefname{figure}{Figure}{Figures}
\crefname{figure}{Fig.}{Figs.}

\def\cvprPaperID{2549} 
\def\confName{CVPR}
\def\confYear{2022}

\newlength\savewidth

\newcommand{\pinknoise}{\texttt{PinkNoise}\xspace} 
\newcommand{\primitives}{\texttt{Primitives}\xspace}
\newcommand{\primitivesS}{\texttt{Primitives-S}\xspace}
\newcommand{\primitivesPS}{\texttt{Primitives-PS}\xspace}

\newcommand{\appnumAblation}{1\xspace}
\newcommand{\appnumConvergeSpeedSynth}{2\xspace}
\newcommand{\appnumQualitativeSynth}{3\xspace}
\newcommand{\appnumConvergeSpeedComp}{4\xspace}
\newcommand{\appnumQualitativeComp}{5\xspace}
\newcommand{\appnumSimilarityFilter}{6\xspace}
\newcommand{\appnumCopyright}{7\xspace}
\newcommand{\appnumCIFAR}{8\xspace}
\newcommand{\appnumPretrainingResult}{9\xspace}
\newcommand{\appnumFFTImage}{10\xspace}
\newcommand{\appnumKMMD}{11\xspace}

\makeatletter
\def\@fnsymbol#1{\ensuremath{\ifcase#1\or \dagger\or \ddagger\or
   \mathsection\or \mathparagraph\or \|\or **\or \dagger\dagger
   \or \ddagger\ddagger \else\@ctrerr\fi}}
\makeatother

\newcommand{\figSynthEx}{
\begin{figure}[t]
\centering
\includegraphics[width=0.49\columnwidth]{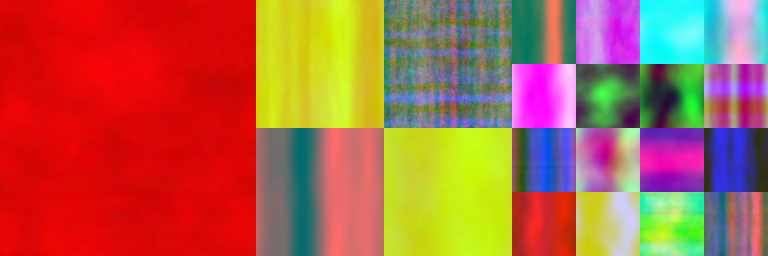}
\includegraphics[width=0.49\columnwidth]{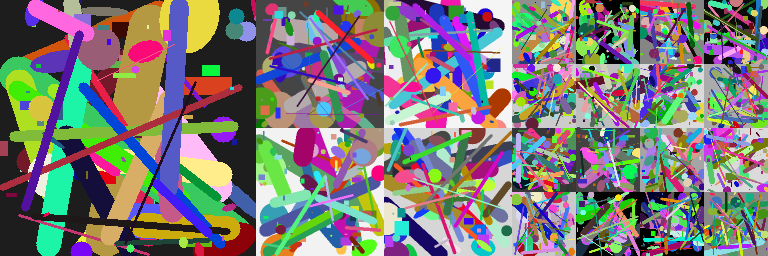} \\
\makebox[0.49\columnwidth][c]{(a) \pinknoise}
\makebox[0.49\columnwidth][c]{(b) \primitives} \\ \vspace{1mm}
\includegraphics[width=0.49\columnwidth]{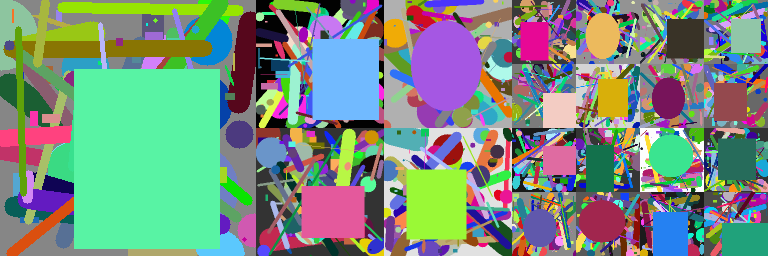}
\includegraphics[width=0.49\columnwidth]{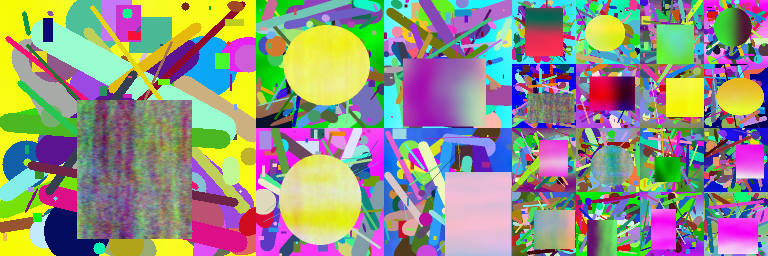} \\
\makebox[0.49\columnwidth][c]{(c) \primitivesS}
\makebox[0.49\columnwidth][c]{(d) \primitivesPS}\\

\caption{Visualization of our synthetic datasets. We visualize four variants of our synthetic datasets and \primitivesPS is finally chosen for the best performance. Example images are resized in three different scales.}
\vspace{-7mm}
\label{fig:synthetic_example}
\end{figure}
}
\newcommand{\figPrimitivesVSPrimitivesPS}{
\begin{figure}[t]
\centering
\includegraphics[width=0.49\columnwidth]{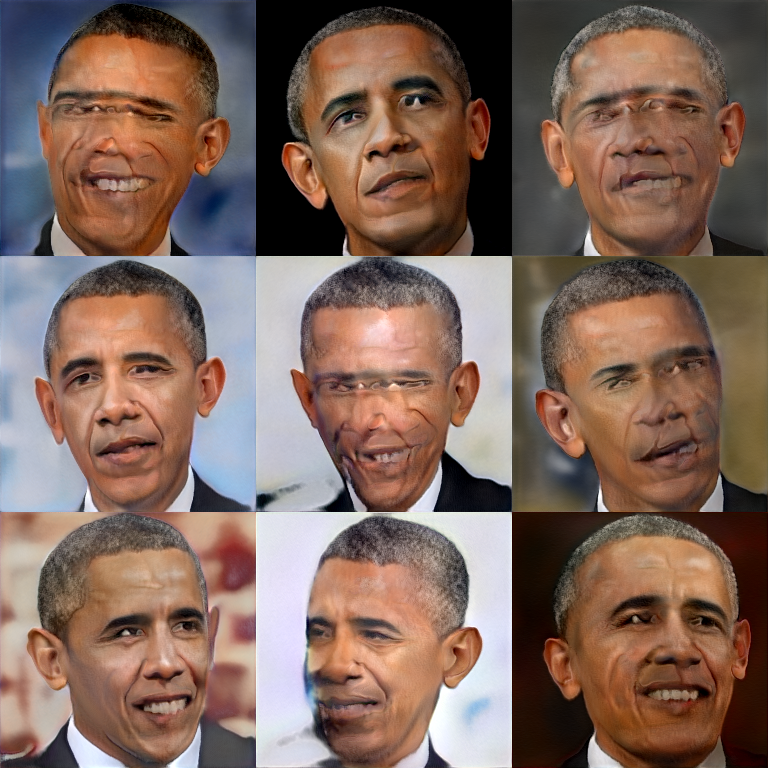}
\includegraphics[width=0.49\columnwidth]{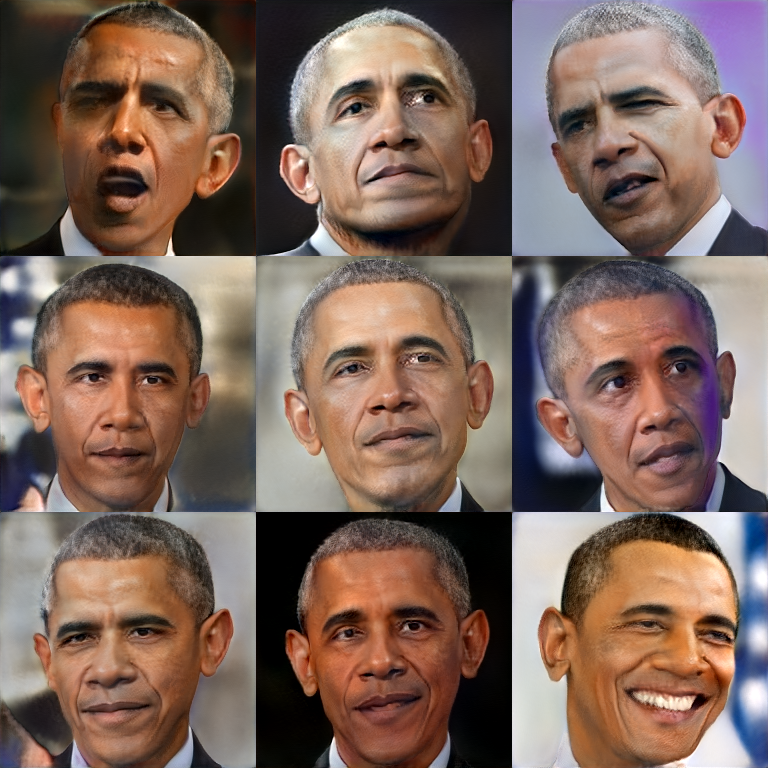} \\
\makebox[0.49\columnwidth][c]{(a)}
\makebox[0.49\columnwidth][c]{(b)}

\caption{Comparison between (a) \primitives and (b) \primitivesPS on Obama dataset. The model pretrained with \primitives generates multiple faces in a single image.}
\label{fig:primitives_vs_primitivesPS}
\vspace{-7mm}
\end{figure}
}
\newcommand{\figAbs}{
\begin{figure}[t]
\centering
\includegraphics[width=0.98\columnwidth]{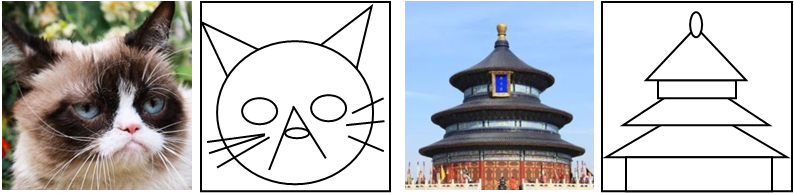} \\

\caption{Potentials of primitive shapes for representing things. We only use a line, ellipse, and rectangle to express a cat and a temple. These examples motivate us to develop \primitives, which generates the data by a simple composition of the shapes.}
\label{fig:abstraction}
\vspace{-2mm}
\end{figure}
}
\newcommand{\figFFT}{
\begin{figure}[h]
\centering
\includegraphics[width=0.32\columnwidth]{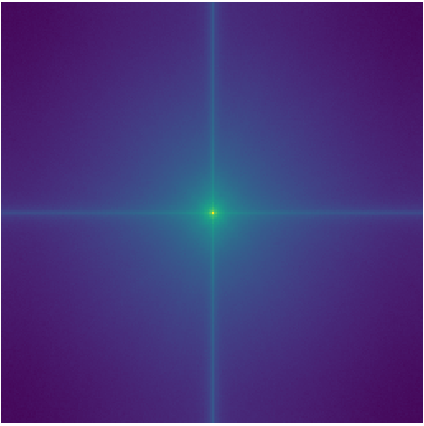}
\includegraphics[width=0.32\columnwidth]{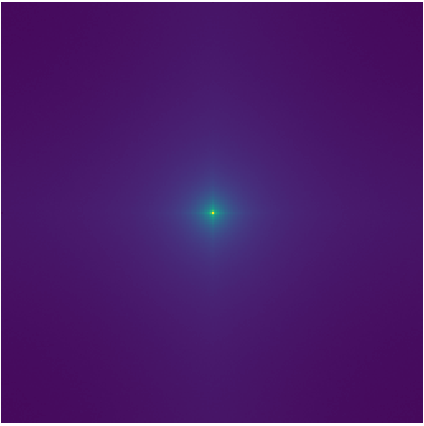}
\includegraphics[width=0.32\columnwidth]{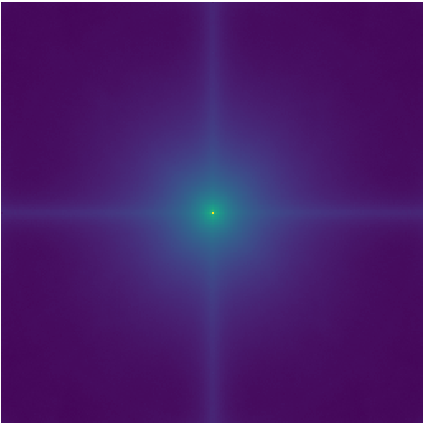} \\
\makebox[0.32\columnwidth][c]{(a)}
\makebox[0.32\columnwidth][c]{(b)}
\makebox[0.32\columnwidth][c]{(c)}

\caption{The magnitude spectrum of (a) Bridge, (b) \pinknoise, and (c) \primitives. We apply FFT on each image and then visualize the average magnitude of the images. When we visualize, we take a logarithmic transformation. Although \pinknoise aims to mimic the magnitude spectrum of natural images, that of \primitives approximates the benchmark dataset better than that of \pinknoise.}
\label{fig:fft}
\end{figure}
}
\newcommand{\figFIDperIT}{
\begin{figure}[t]
\centering
\includegraphics[width=0.99\columnwidth]{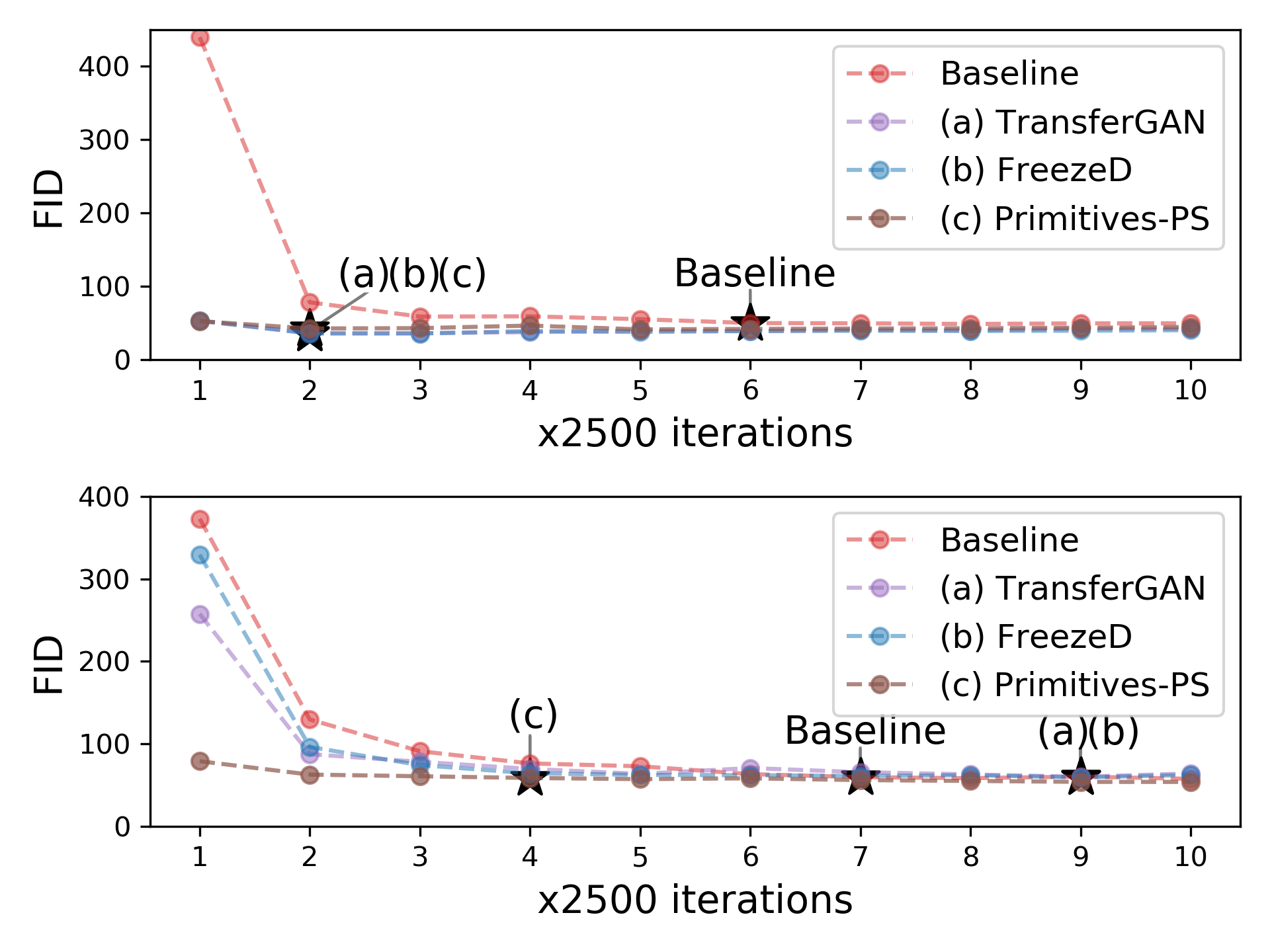} \\
\vspace{-3mm}
\caption{FID per training iterations. The star marker ($\bigstar$) indicates the point where the model reaches 95\% of the best FID score of the from scratch model with DiffAug (baseline). Our \primitivesPS pretrained model is comparable to the competitors on Obama dataset (upper) and converges faster than the others on Bridge of sighs dataset (lower).}
\label{fig:fid_per_it}
\vspace{-5mm}
\end{figure}
}
\newcommand{\figDiff}{
\begin{figure}[t]
\centering
\includegraphics[width=0.99\columnwidth]{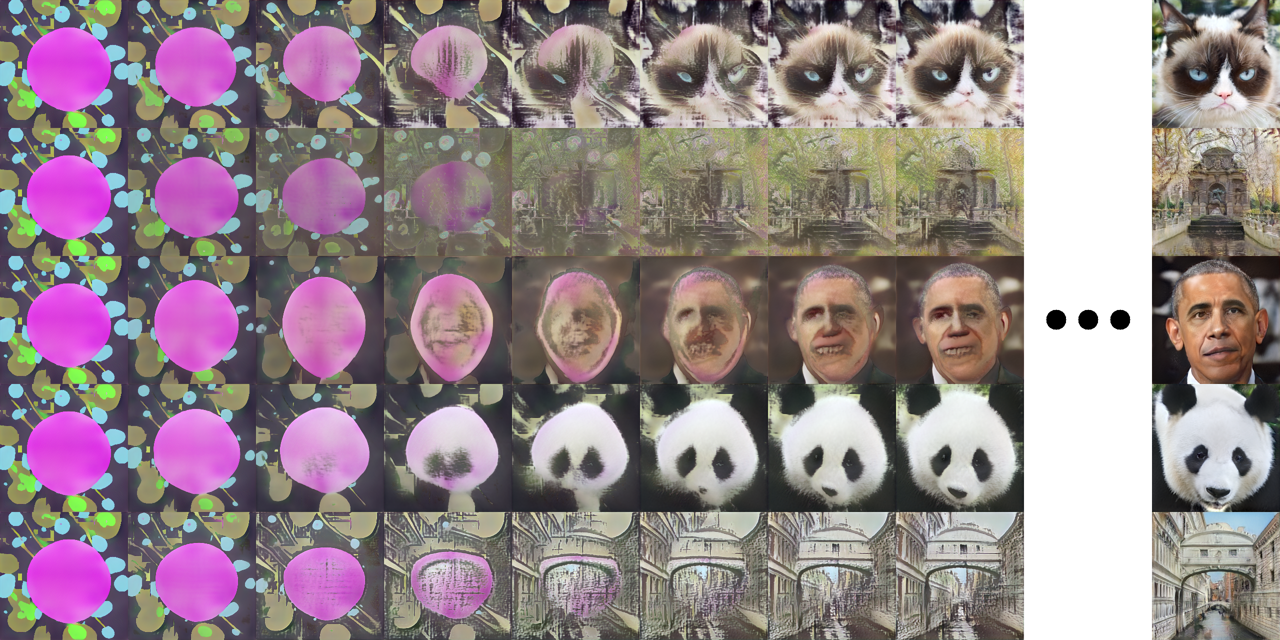} \\

\caption{Morphing upon the transfer learning iterations of the \primitivesPS pretrained model. We generate the images by using the same latent vector. The center lilac circles are gradually changed into salient regions.}
\label{fig:differentiation}
\vspace{-4mm}
\end{figure}
}
\newcommand{\figLowShotQual}{
\begin{figure*}[t]
\centering
\includegraphics[width=0.245\textwidth]{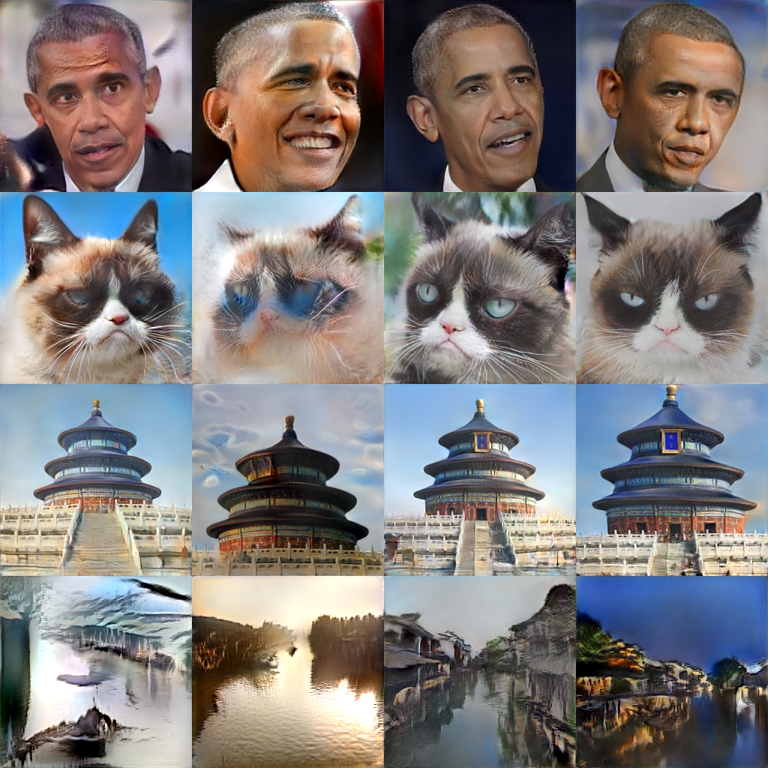}
\includegraphics[width=0.245\textwidth]{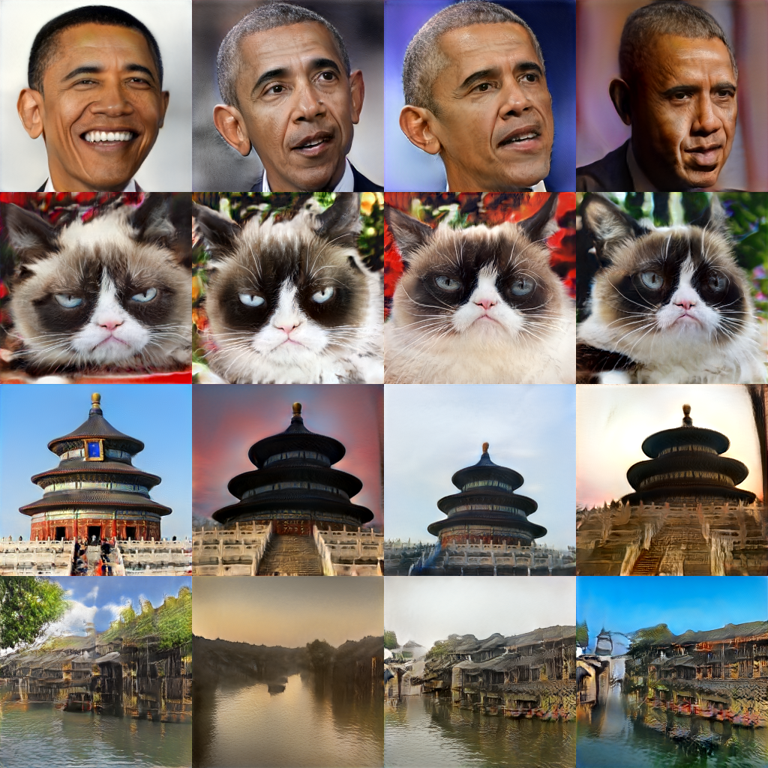}
\includegraphics[width=0.245\textwidth]{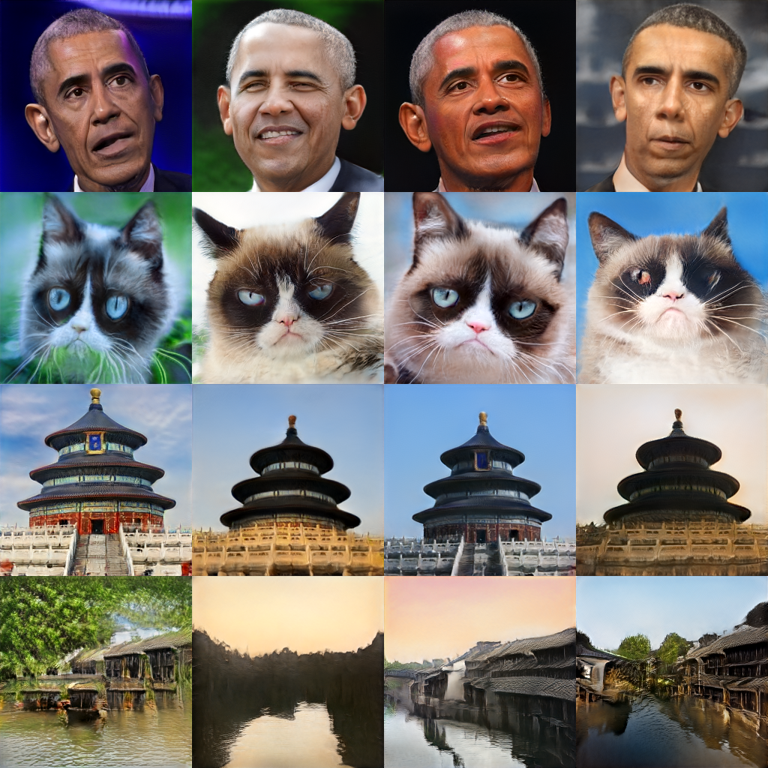}
\includegraphics[width=0.245\textwidth]{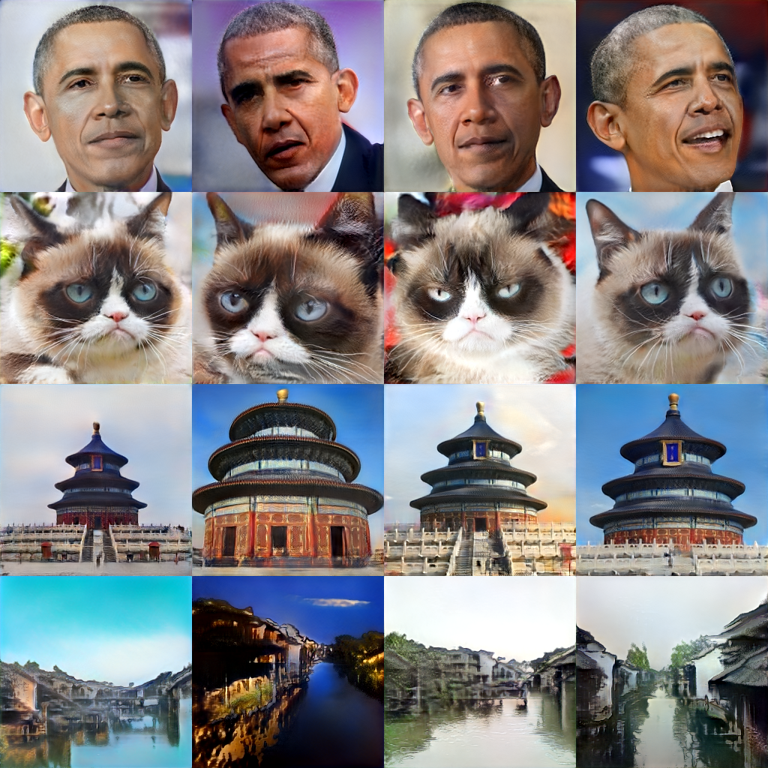} \\
\makebox[0.245\textwidth][c]{(a) From scratch}
\makebox[0.245\textwidth][c]{(b) TransferGAN}
\makebox[0.245\textwidth][c]{(c) FreezeD}
\makebox[0.245\textwidth][c]{(d) \primitivesPS}
\vspace{-2mm}
\caption{Qualitative evaluation on Obama, Grumpy cat, Temple, and Wuzhen. For more results, please refer to the supplementary \appnumQualitativeComp.}
\label{fig:lowshot_qual}
\vspace{-1mm}
\end{figure*}
}
\newcommand{\figCIFAR}{
\begin{figure*}[t]
\centering
\includegraphics[width=0.33\textwidth]{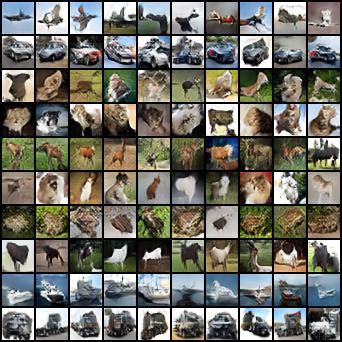}
\includegraphics[width=0.33\textwidth]{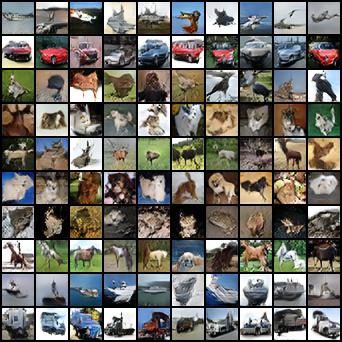}
\includegraphics[width=0.33\textwidth]{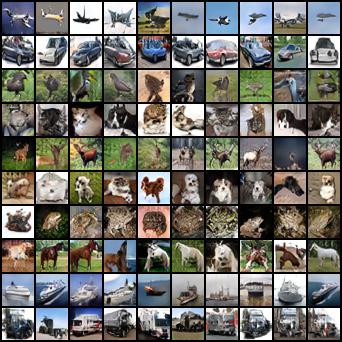} \\
\makebox[0.33\textwidth][c]{(a) From scratch}
\makebox[0.33\textwidth][c]{(b) + DiffAug}
\makebox[0.33\textwidth][c]{(c) + \primitivesPS pretraining}
\vspace{-6mm}
\caption{Qualitative evaluation on CIFAR-10 dataset with 10\% of samples. Each row contains samples in the same class.}
\vspace{-4mm}
\label{fig:cifar}
\end{figure*}
}
\newcommand{\figFIDperITSynthAppendix}{
\begin{figure*}[t!]
\centering
\vspace{-10mm}
\includegraphics[width=0.45\textwidth]{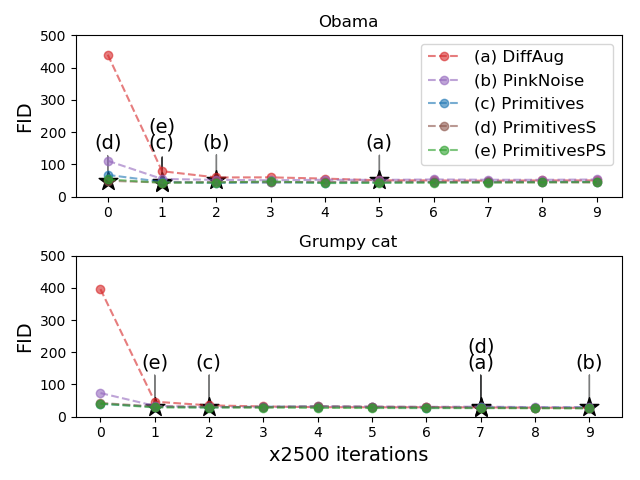} 
\includegraphics[width=0.45\textwidth]{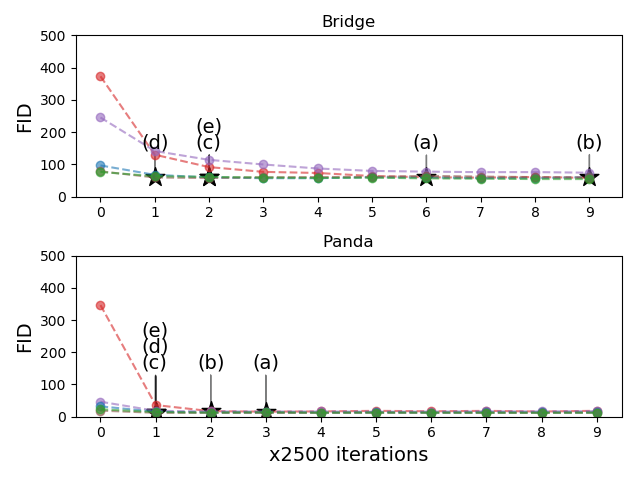}
\vspace{-3mm}
\caption{FID per training iterations. The star marker ($\bigstar$) indicates the point where the model reaches 95\% of the best FID score of the from scratch model with DiffAug (baseline). The legend is the same for all graphs.}
\label{fig:fid_per_it_app}
\end{figure*}
}
\newcommand{\figFIDperITCompAppendix}{
\begin{figure*}[t]
\centering
\vspace{-10mm}
\includegraphics[width=0.45\textwidth]{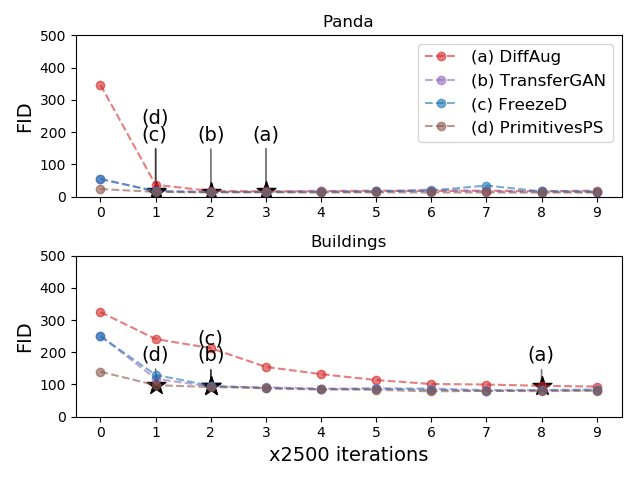} 
\includegraphics[width=0.45\textwidth]{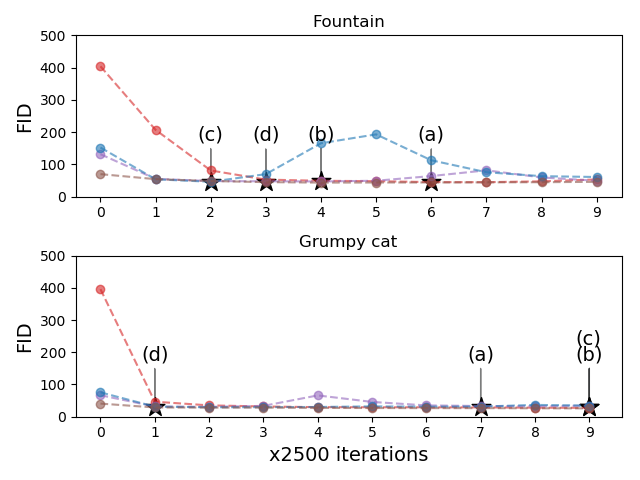} \\
\includegraphics[width=0.45\textwidth]{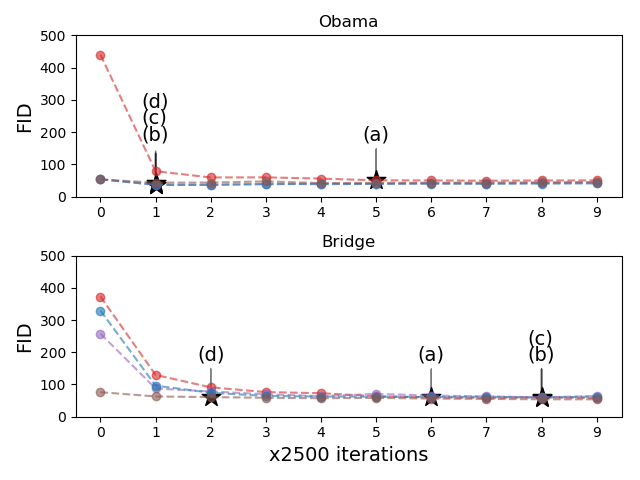} 
\includegraphics[width=0.45\textwidth]{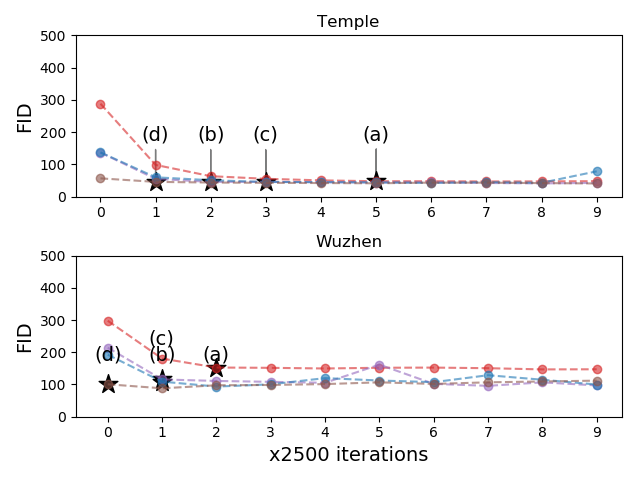} 
\vspace{-3mm}
\caption{The additional results of Figure 6 in the main text. FID per training iterations. The star marker ($\bigstar$) indicates the point where the model reaches 95\% of the best FID score of the from scratch model with DiffAug (baseline). The legend is the same for all graphs.}
\label{fig:fid_per_it_comp_app}
\end{figure*}
}
\newcommand{\figAppSynthQualPNandPrimitives}{
\begin{figure*}[t]
\centering
\includegraphics[width=0.99\textwidth]{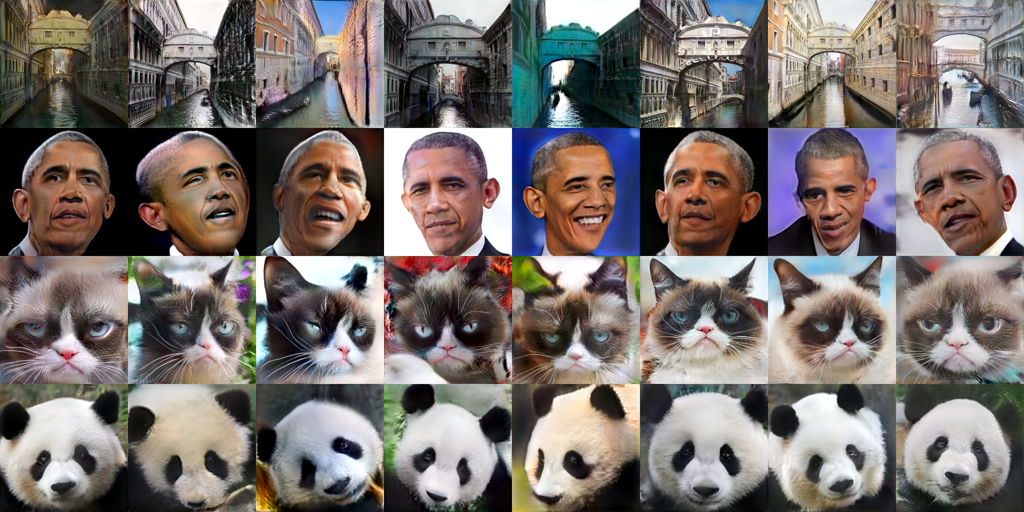} \\
\makebox[0.99\textwidth][c]{(a) \pinknoise} \vspace{2mm}
\includegraphics[width=0.99\textwidth]{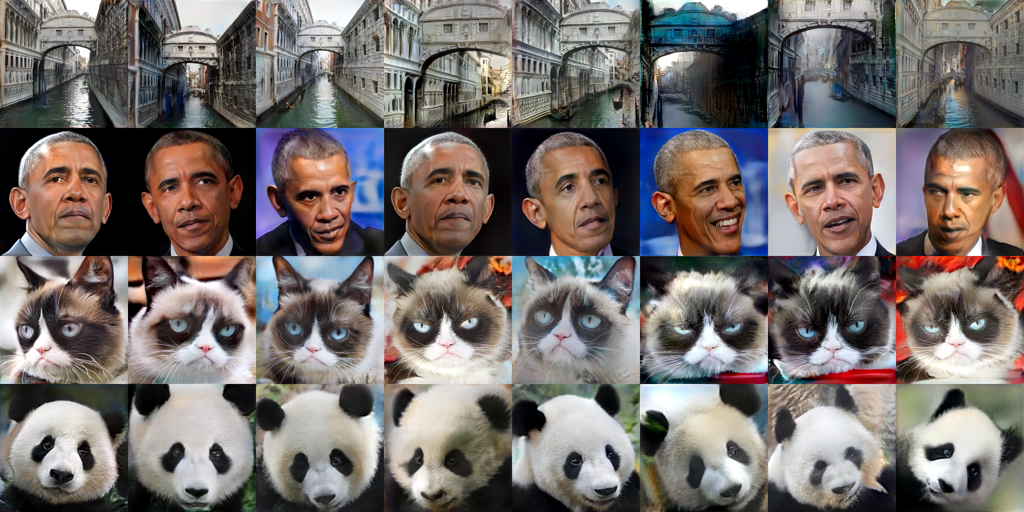} \\
\makebox[0.99\textwidth][c]{(b) \primitives}
\caption{Low-shot image generation results of the models transferred from \pinknoise and \primitives.}
\label{fig:app_qual_pinknoise_primitives}
\end{figure*}
}
\newcommand{\figAppSynthQualSandPS}{
\begin{figure*}[t]
\centering
\includegraphics[width=0.99\textwidth]{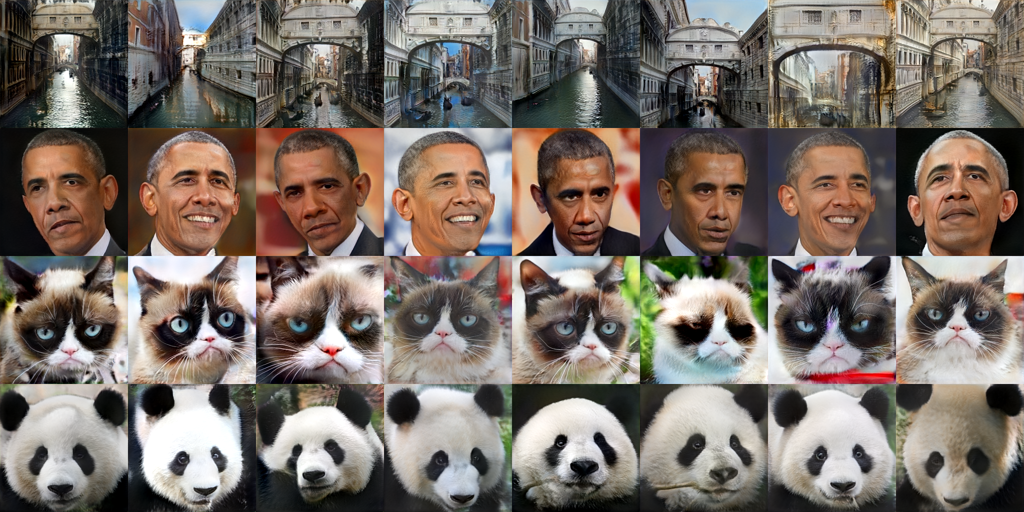} \\
\makebox[0.99\textwidth][c]{(a) \primitivesS} \vspace{2mm}
\includegraphics[width=0.99\textwidth]{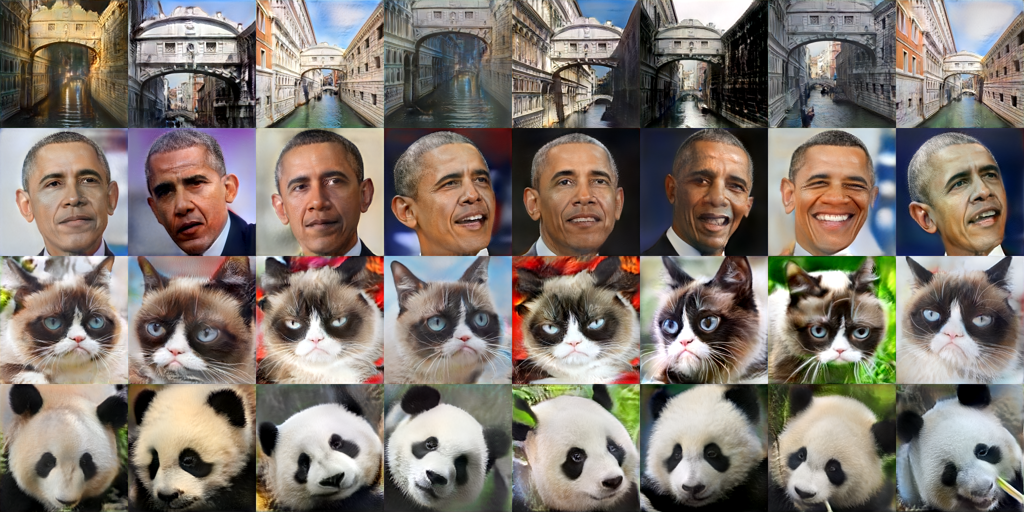} \\
\makebox[0.99\textwidth][c]{(b) \primitivesPS}
\caption{Low-shot image generation results of the models transferred from \primitivesS and \primitivesPS.}
\label{fig:app_qual_S_PS}
\end{figure*}
}
\newcommand{\figAppQualDiffAug}{
\begin{figure*}[b]
\centering
\includegraphics[width=0.99\textwidth]{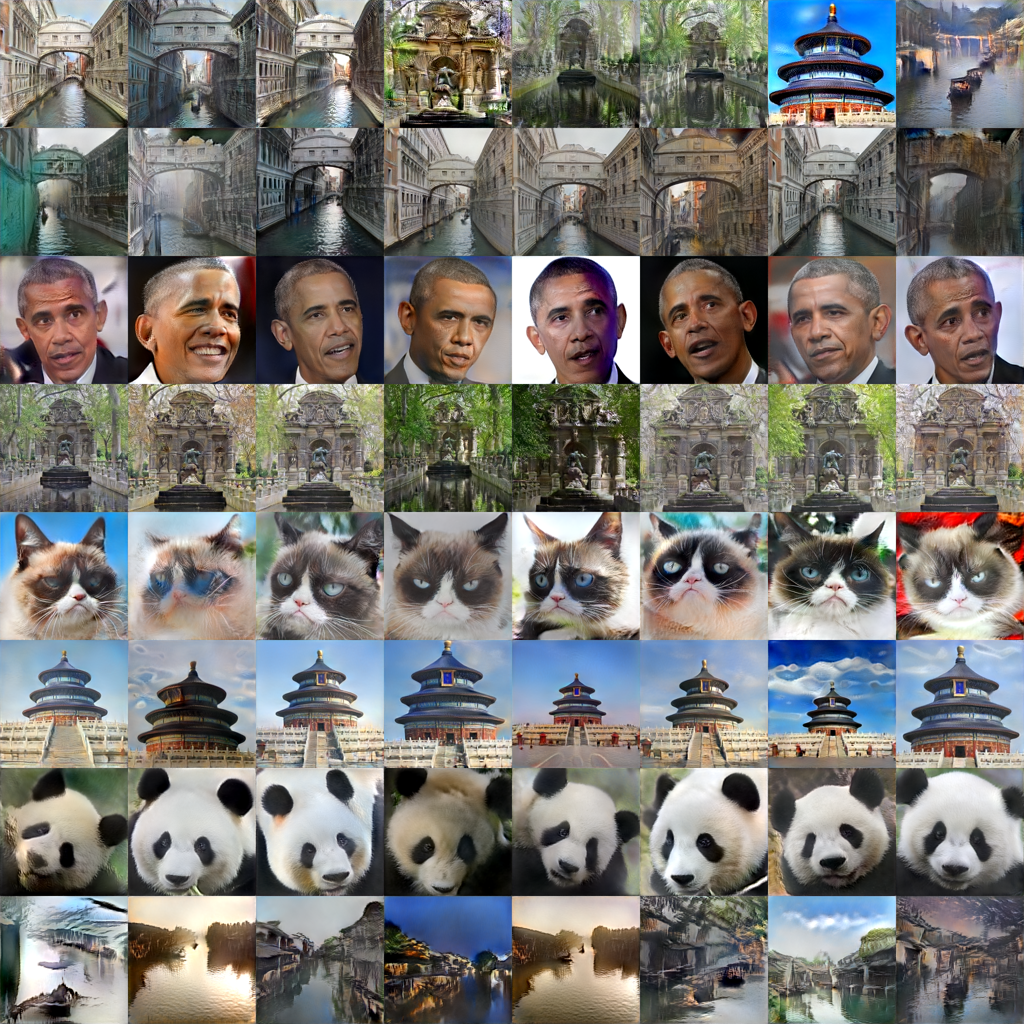} 
\caption{The additional generated samples of Figure 5 in the main text. The images are generated with the model trained from scratch.}
\label{fig:app_qual_diffaug}
\end{figure*}
}
\newcommand{\figAppQualTransferGAN}{
\begin{figure*}[t]
\centering
\includegraphics[width=0.99\textwidth]{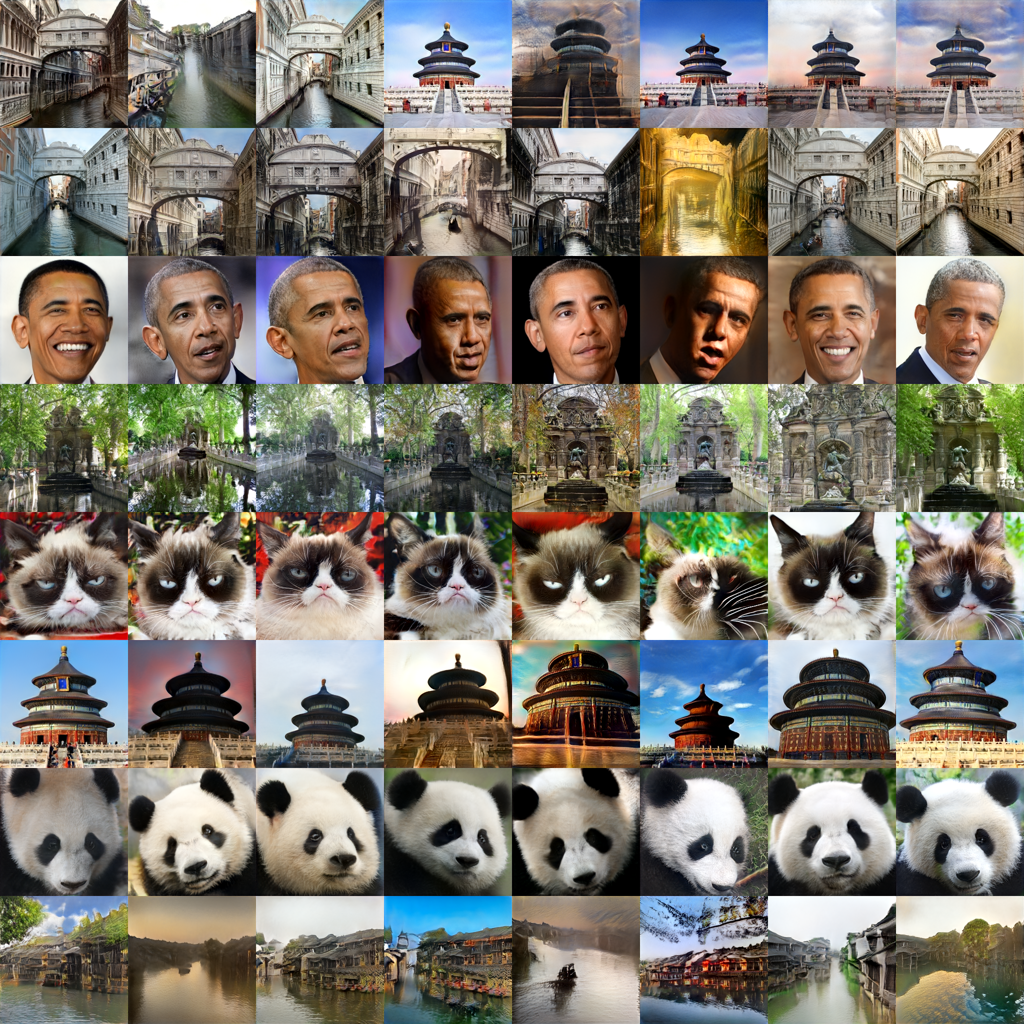} 
\vspace{-3mm}
\caption{The additional generated samples of Figure 5 in the main text. The images are generated with the model pretrained with FFHQ and transferred by using TransferGAN.}
\label{fig:app_qual_transfergan}
\end{figure*}
}
\newcommand{\figAppQualFreezeD}{
\begin{figure*}[t]
\centering
\includegraphics[width=0.99\textwidth]{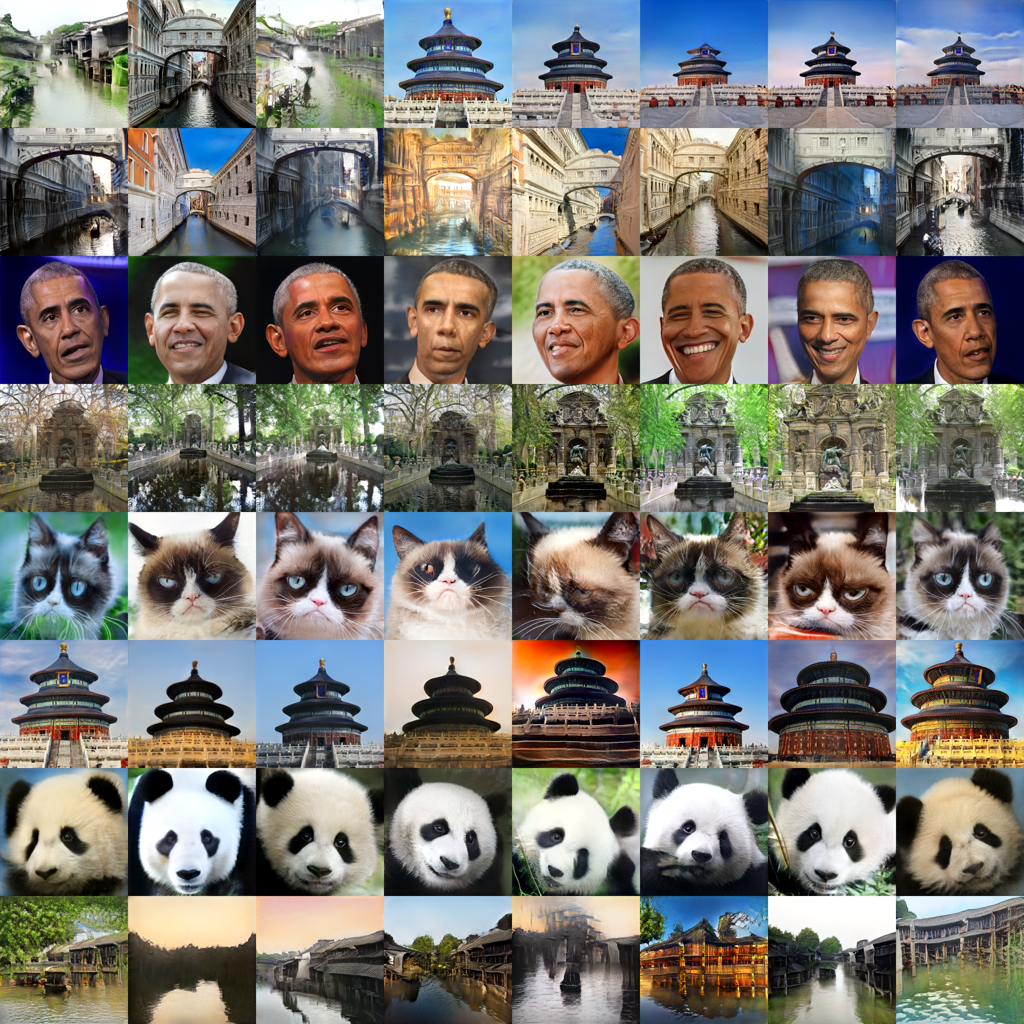} 
\vspace{-3mm}
\caption{The additional generated samples of Figure 5 in the main text. The images are generated with the model pretrained with FFHQ and transferred by using FreezeD.}
\label{fig:app_qual_freezed}
\end{figure*}
}

\newcommand{\figAppQualOurs}{
\begin{figure*}[t]
\centering
\includegraphics[width=0.99\textwidth]{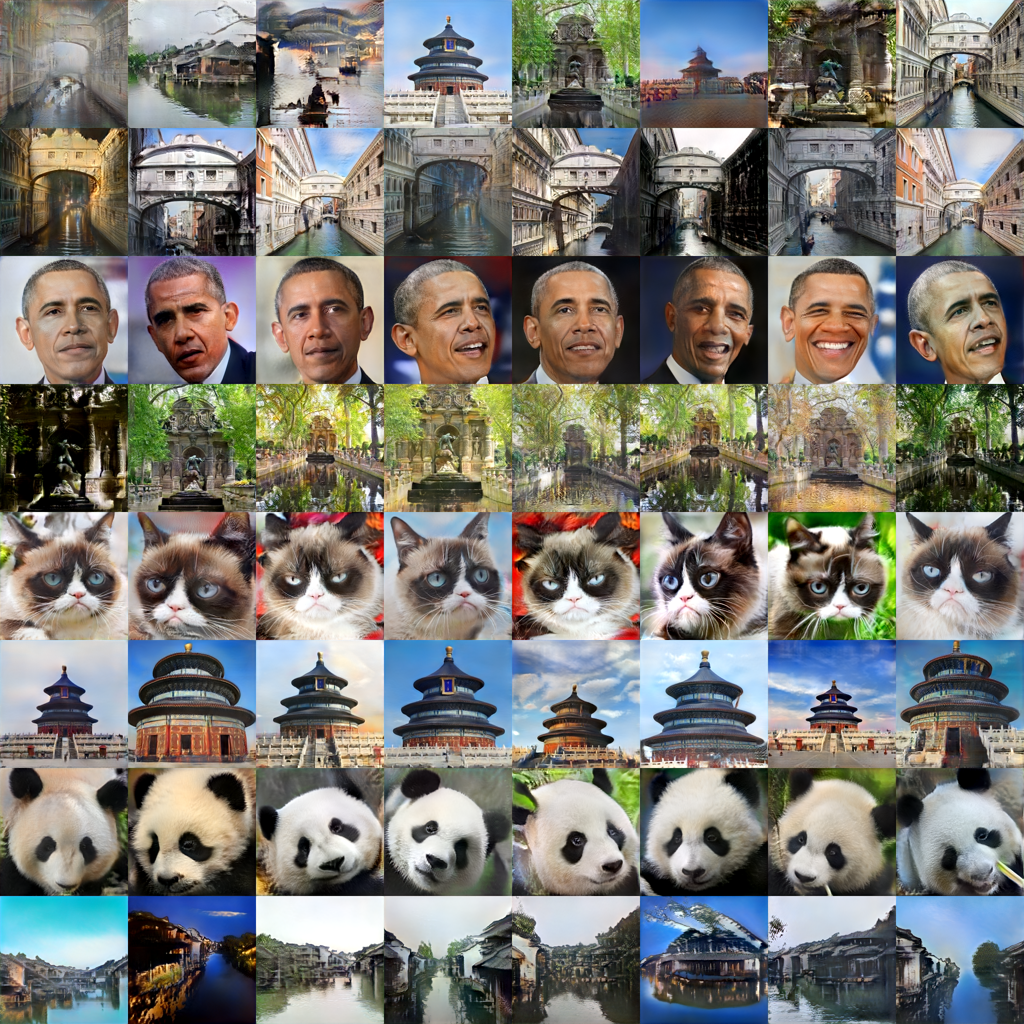} 
\caption{The additional generated samples of Figure 5 in the main text. The images are generated with the model pretrained with our \primitivesPS.}
\label{fig:app_qual_ours}
\end{figure*}
}
\newcommand{\figAppCifarLeak}{
\begin{figure}[h]
\centering
\includegraphics[width=0.99\columnwidth]{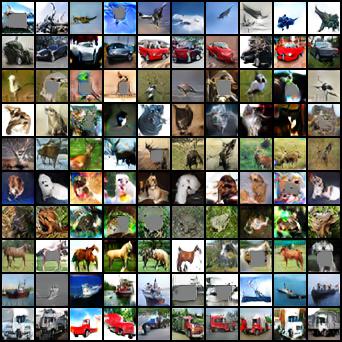} 
\caption{Examples of the leakage when using DiffAug. The gray box in some images shows the leakage of cutout operation.}
\label{fig:app_cifar_leak}
\end{figure}
}
\newcommand{\figAppCifarOurs}{
\begin{figure}[h]
\centering
\includegraphics[width=0.99\columnwidth]{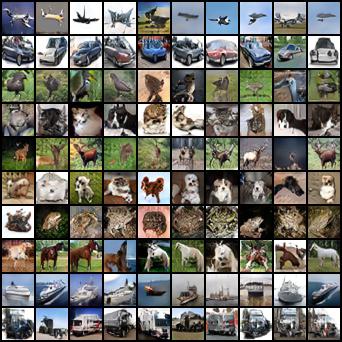} 
\caption{Outputs of the model transferred from our model on CIFAR-10. The model does not suffer from augmentation leakage although we use DiffAug.}
\label{fig:app_cifar_ours}
\end{figure}
}
\newcommand{\figAppPrimitivesPSFakes}{
\begin{figure}[h]
\centering
\includegraphics[width=0.99\columnwidth]{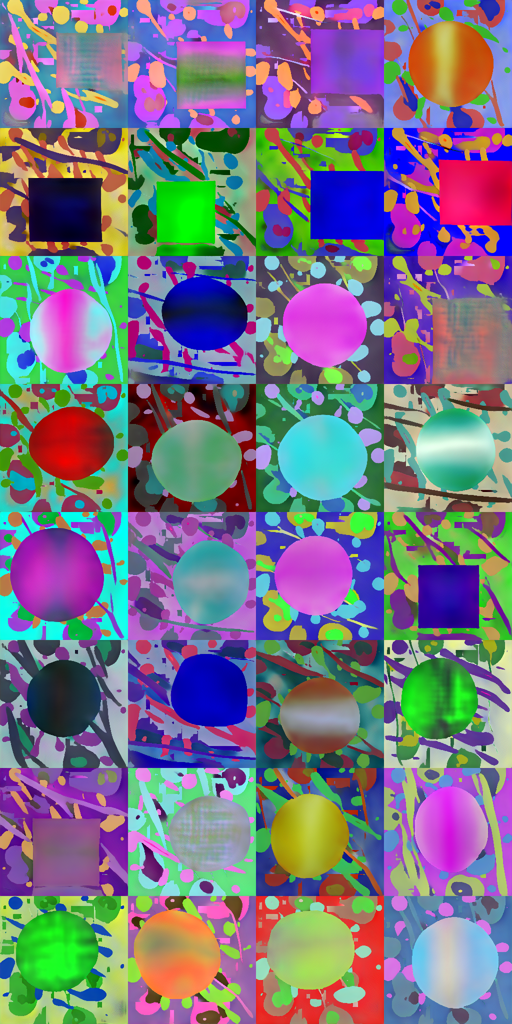} 
\caption{The outputs of the model pretrained with \primitivesPS. The generated outputs are similar to the synthetic samples.}
\label{fig:app_primitivesPS_fakes}
\end{figure}
}

\newcommand{\tabMembership}{
\begin{table}[t!]
    \centering
    \caption{Membership inference performance on the source dataset by attacking a transferred classifier as reported in \cite{zou2020privacy}.
    }
    \resizebox{0.8\columnwidth}{!}{
    \begin{tabular}{c c c c c}
            \hline
            Dataset & AUC & Accuracy & Precision & Recall\\ \hline
            CIFAR100 & 0.522 & 0.502 & 0.478 & 0.523\\
            Flowers102 & 0.528 & 0.496 & 0.432 & 0.505 \\
            PubFig83 & 0.495 & 0.481 & 0.396 & 0.524 \\ 
            \hline\\
    \end{tabular}
    }
  \label{tab:membership}
\end{table}
}
\newcommand{\tabCompareSynth}{
\begin{table}[t!]
    \centering
    \caption{The FID score of transferring to low-shot datasets from the proposed pretraining datasets. The lower is the better. Bold and underlined text indicates the best and second best performance among the pretraining datasets. It will be the same convention throughout the paper.}
    \resizebox{0.9\columnwidth}{!}{
    \begin{tabular}{l c c c c}
            \hline
            \diagbox[innerwidth=\textwidth*1/8]{Source}{Target} & Obama & Grumpy cat & Bridge & Panda \\ \hline
            Scratch + DiffAug & 48.98 & 27.51 & 57.72 & 15.82 \\
            \pinknoise & 50.32 & 29.47 & 73.82 & 15.65 \\
            \primitives & \underline{43.20} & 27.97 & 59.89 & 12.78 \\
            \primitivesS & 43.29  & \underline{26.57} & \underline{57.24} & \textbf{11.95} \\
            \primitivesPS & \textbf{41.62} & \textbf{26.01} & \textbf{54.02} & \underline{12.23} \\
            \hline\\
    \end{tabular}
    }
  \label{tab:compare_synth}
  \vspace{-7mm}
\end{table}
}
\newcommand{\tabFreqDist}{
\begin{table}[t!]
    \centering
    \caption{SSIM between the magnitude spectrum of the frequency domain of the synthetic and target dataset. The higher score means the more similar pair. We observed that the tendency is the same with L1 or L2 distance.}
    \resizebox{0.99\columnwidth}{!}{
    \begin{tabular}{l c c c c c |c}
            \hline
            \diagbox[innerwidth=\textwidth*1/8]{Source}{Target} & Obama & Grumpy cat & Bridge & Panda & FFHQ & Mean\\ \hline
            \pinknoise & 0.8368 & 0.8148 & 0.7676 & 0.8328 & 0.8553 & 0.8215\\
            \primitives & 0.9309 & 0.9366 & 0.9198 & 0.9200 & 0.9635 & 0.9342\\
            \primitivesS & \underline{0.9421} & \underline{0.9463} & \textbf{0.9308} & \underline{0.9334} & \underline{0.9756} & \underline{0.9456}\\
            \primitivesPS & \textbf{0.9432} & \textbf{0.9476} & \underline{0.9307}& \textbf{0.9352} & \textbf{0.9767} & \textbf{0.9467}\\
            \hline\\
    \end{tabular}
    }
    \vspace{-5mm}
  \label{tab:freq_distance}
\end{table}
}
\newcommand{\tabCompareTransfer}{
\begin{table*}[t!]
    \centering
    \caption{The FID score of transferred models to low-shot datasets. We use FFHQ pretrained weight for TransferGAN and FreezeD. For all models, we apply DiffAug. Bold and underlined text indicates the best and second best performance among the pretraining datasets.}
    \vspace{-2mm}
    \resizebox{0.85\textwidth}{!}{
    \begin{tabular}{l c c c c c c c c}
            \hline
            \diagbox[innerwidth=\textwidth*1/9]{Source}{Target} & Obama & Grumpy cat & Bridge & Panda & Temple & Wuzhen & Fountain & Buildings \\ \hline
            Scratch + DiffAug~\cite{zhao2020diffaugment} & 48.98 & \underline{27.51} & \underline{57.72} & 15.82 & 46.69 & 146.81 & \underline{44.46} & 93.71 \\
            TransferGAN~\cite{wang2018transfergan} & \underline{36.50} & 30.60 & 60.29 & 14.53 & \underline{40.58} & 95.83 & 46.61 & 81.63 \\
            FreezeD~\cite{mo2020freezed} & \textbf{35.90} & 29.41 & 59.47 & \underline{13.39} & 42.09 & \underline{93.54} & 45.70 & \underline{80.48} \\
            \primitivesPS & 41.62 & \textbf{26.01} & \textbf{54.02} & \textbf{12.23} & \textbf{40.42} & \textbf{88.14} & \textbf{43.06} & \textbf{78.74} \\
            \hline\\
    \end{tabular}
    }
  \label{tab:compare_transfer}
  \vspace{-7mm}
\end{table*}
}
\newcommand{\tabFilterDiv}{
\begin{table}[t!]
    \centering
    \caption{The average consine similarity between the filters in the same layer. The lower value indicates the more diverse filters.}
    \vspace{-2mm}
    \resizebox{0.75\columnwidth}{!}{
    \begin{tabular}{c c c}
            \hline
            Pretraining DB & Discriminator & Generator \\ \hline
            \primitivesPS & \textbf{0.00820} & \textbf{0.00828}\\
            FFHQ & 0.01348 & 0.01434 \\
            \hline\\
    \end{tabular}
    }
  \label{tab:filter_diversity}
  \vspace{-4mm}
\end{table}
}
\newcommand{\tabFilterDivApp}{
\begin{table}[h]
    \centering
    \caption{The additional results of Table 4 in the main text. The average consine similarity between the filters in the same layer. The lower value indicates the more diverse set of filters.}
    \resizebox{0.98\columnwidth}{!}{
    \begin{tabular}{c c c| c c}
            \hline
            & \multicolumn{2}{c}{Discriminator} & \multicolumn{2}{c}{Generator} \\ \hline
            & \primitivesPS & FFHQ & \primitivesPS & FFHQ \\ \hline
            conv0  & \textbf{0.00660} & 0.01245          & \textbf{0.00315} & 0.00685          \\
            conv1  & 0.02104          & \textbf{0.00932} & \textbf{0.00273} & 0.00843          \\
            conv2  & 0.01012          & \textbf{0.00779} & \textbf{0.00291} & 0.00956          \\
            conv3  & \textbf{0.00839} & 0.01216          & \textbf{0.00348} & 0.01080          \\
            conv4  & \textbf{0.00607} & 0.00713          & \textbf{0.00539} & 0.01059          \\
            conv5  & \textbf{0.00596} & 0.00668          & \textbf{0.00329} & 0.01406          \\
            conv6  & \textbf{0.00507} & 0.00563          & \textbf{0.00363} & 0.01199          \\
            conv7  & \textbf{0.00632} & 0.00714          & \textbf{0.00433} & 0.01465          \\
            conv8  & 0.00380          & \textbf{0.00365} & \textbf{0.00652} & 0.01317          \\
            conv9  & \textbf{0.00521} & 0.00703          & \textbf{0.00933} & 0.01626          \\
            conv10 & 0.00503          & \textbf{0.00420} & \textbf{0.01133} & 0.01778          \\
            conv11 & \textbf{0.00462} & 0.00760          & 0.01981          & \textbf{0.01977} \\
            conv12 & \textbf{0.01844} & 0.08438          & \textbf{0.03176} & 0.03250          \\ \hline
            Mean   & \textbf{0.00820} & 0.01348          & \textbf{0.00828} & 0.01434          \\
            \hline\\
    \end{tabular}
    }
  \label{tab:filter_diversity_app}
\end{table}
}
\newcommand{\tabCIFAR}{
\begin{table}[t!]
    \centering
    \caption{The FID of BigGAN, with DiffAug, and with DiffAug initialized by \primitivesPS (\texttt{PS}) pretrained model on CIFAR. '*' indicates the best FID before augmentation leakage~\cite{Neurips2020Ada}. Please refer to the supplementary \appnumCIFAR for the details.}
    \vspace{-2mm}
    \resizebox{0.99\columnwidth}{!}{
    \begin{tabular}{l c c c| c c c}
            \hline
             & \multicolumn{3}{c}{CIFAR-10} & \multicolumn{3}{c}{CIFAR-100}\\ \hline
             & 10\% & 20\% & 100\% & 10\% & 20\% & 100\% \\ \hline
BigGAN & 44.14                & 20.80                 & 9.45                 & 66.21                & 34.78                & 13.45                \\
+ DiffAug & 29.78*               & 14.04                & \textbf{8.55}        & 41.70*                & 21.14                & 11.51                \\
+ Pretrained (\texttt{PS}) & \textbf{21.33}       & \textbf{12.79}       & 8.79                 & \textbf{32.57}       & \textbf{20.58}       & \textbf{11.29}       \\
            \hline\\
    \end{tabular}
    }
  \label{tab:cifar}
\vspace{-7mm}
\end{table}
}

\newcommand{\tabAblationMerge}{
\begin{table}[t!]
    \centering
    \caption{Ablation study on the policy to determine the size of each particle (upper) and the number of particles (lower).}
        \vspace{-2mm}
    \resizebox{0.98\columnwidth}{!}{
    \begin{tabular}{l c c c c}
            \hline
            Policy & Obama & Grumpy cat & Bridge & Panda \\ \hline
\textbf{Fix} ($\sfrac{1}{10}$) & 48.30           & 29.74          & 63.00             & 17.69          \\
\textbf{Fix} ($\sfrac{1}{5}$) &46.41          & 29.22          & 64.02          & 14.97          \\
\textbf{Fix} ($\sfrac{1}{2}$) &48.05          & 29.37          & 64.65          & 15.14          \\
\footnotesize{\pinknoise + \texttt{PS}} & 49.13 & 29.87 & 66.00 & 15.12 \\
\textbf{Rand} &44.85          & 29.84          & 60.45          & 14.67          \\
\textbf{Decay} &\textbf{41.62} & \textbf{26.01} & \textbf{54.02} & \textbf{12.23} \\
            \hline
\# of particles & Obama          & Grumpy cat            & Bridge           & Panda          \\ \hline
0            & 49.13          & 29.87          & 66.00             & 15.12          \\
10           & 44.10           & 28.00             & 63.26          & 13.35          \\
50           & 42.49          & 28.40           & 59.17          & \textbf{11.79} \\
100          & \textbf{41.62} & \textbf{26.01} & 54.02          & 12.23          \\
500          & 42.45          & 27.92          & \textbf{52.27} & 12.12          \\
            \hline\\
    \end{tabular}
    }
    \vspace{-10mm}
  \label{tab:ablation_merge}
\end{table}
}
\newcommand{\tabKMMDApp}{
\begin{table}[t!]
    \centering
    \vspace{-2mm}
        \caption{\footnotesize KMMD score for \textbf{Table 3 in the main text (256)}.}
    \vspace{-3.5mm}
    \resizebox{\columnwidth}{!}{
    \begin{tabular}{l c c c c c c c c}
            \hline
             & Obama & Cat & Brid. & Panda & Temp. & Wuzhen & Fountain & Build. \\ \hline
            DfAug & {0.23} & {0.15} & {0.23} & {0.28} & {0.18} & {0.39} & {0.21} & {0.21} \\
            TGAN & {0.13} & {0.14} & {0.22} & {0.21} & {0.14} & {0.27} & {0.19} & {0.18} \\ 
            FrzD & \textbf{{0.12}} & \textbf{{0.14}} & {0.22} & \textbf{{0.18}} & \textbf{{0.13}} & {0.25} & {0.21} & \textbf{{0.16}} \\ 
            Ours & {0.17} & {0.15} & \textbf{{0.17}} & {0.26} & {0.14} & \textbf{{0.25}} & \textbf{{0.17}} & {0.18} \\ 
            \hline
    \end{tabular}
    }
  \label{tab:KMMD}
  \vspace{-7mm}
\end{table}
}
\newcommand{\tabImageNetApp}{
\begin{table}[t!]
    \centering
    \vspace{-2mm}
        \caption{\footnotesize {FID} score of \textit{ImageNet pretrained} model and \primitivesPS pretrained model on 512$\times$512.}
    \vspace{-3.5mm}
    \resizebox{\columnwidth}{!}{
    \begin{tabular}{l c c c c c c c c}
            \hline
             & Obama & Cat & Brid. & Panda & Temp. & Wuzhen & Fountain & Build. \\ \hline
            DfAug & {59.6} & {\underline{28.0}} & {147.8} & {14.4} & {45.0} & {150.9} & {214.2} & {99.2} \\
             
            TGAN & \textbf{{37.5}} & {35.2} & {52.0} & {\underline{11.8}} & {42.5} & {84.1} & {284.3} & {65.5} \\
             
            FrzD & {\underline{39.1}} & {28.8} & \textbf{{48.6}} & \textbf{{11.2}} & \textbf{{38.9}} & \textbf{{69.5}} & \textbf{{34.3}} & \textbf{{60.2}} \\
             
            Ours & {50.8} & \textbf{{27.7}} & {\underline{51.6}} & {14.9} & {\underline{41.9}} & {\underline{81.6}} & {\underline{42.9}} & {\underline{80.9}} \\
            \hline
    \end{tabular}
    }
  \label{tab:ImageNet}
\end{table}
}

\begin{document}


\title{Commonality in Natural Images Rescues GANs:\\ Pretraining GANs with Generic and Privacy-free Synthetic Data}


\author{Kyungjune Baek, \ Hyunjung Shim\thanks{Hyunjung Shim is a corresponding author.}\\
Yonsei University\\
{\tt\small \{bkjbkj12, kateshim\}@yonsei.ac.kr}
}
\maketitle

\begin{abstract}
Transfer learning for GANs successfully improves generation performance under low-shot regimes. However, existing studies show that the pretrained model using a single benchmark dataset is not generalized to various target datasets. More importantly, the pretrained model can be vulnerable to copyright or privacy risks as membership inference attack advances. To resolve both issues, we propose an effective and unbiased data synthesizer, namely \primitivesPS, inspired by the generic characteristics of natural images. Specifically, we utilize 1) the generic statistics on the frequency magnitude spectrum, 2) the elementary shape (i.e., image composition via elementary shapes) for representing the structure information, and 3) the existence of saliency as prior. Since our synthesizer only considers the generic properties of natural images, the single model pretrained on our dataset can be consistently transferred to various target datasets, and even outperforms the previous methods pretrained with the natural images in terms of Fr\'echet inception distance. Extensive analysis, ablation study, and evaluations demonstrate that each component of our data synthesizer is effective, and provide insights on the desirable nature of the pretrained model for the transferability of GANs.
\end{abstract}
\section{Introduction}
\label{sec:intro}
Generative adversarial networks (GANs)~\cite{2014GANs} are a powerful generative model that can synthesize complex data by learning the implicit density distribution with adversarial training. Thanks to the impressive generation quality, particularly in image generation tasks~\cite{mescheder2018R1reg,karras2020StyleGAN2,brock2018biggan}, GANs have been widely used in various downstream tasks in computer vision, such as data augmentation~\cite{choe2017face}, super-resolution~\cite{ledig2017SRGAN,zhang2019ranksrgan}, image translation~\cite{choi2020starganv2,baek2021tunit}, and image synthesis with primitive representation~\cite{li2018layoutgan,park2019spade}. Despite the remarkable quality, GANs require at least several thousand, mostly several hundred thousand images for training. This requirement for data collection is often infeasible in practical applications (e.g., many pictures of a treasure, endangered species, or the medical images of rare disease). 

The idea of transfer learning has been recently introduced to GAN training~\cite{wang2018transfergan,mo2020freezed} for resolving the real-world generation problem. Following the common practice, the general framework of GAN transfer learning 1) pretrains GANs on a publicly available large-scale source dataset (e.g., FFHQ and ImageNet) and then 2) finetunes GANs with a relatively small target dataset. As a result, developing GANs with transfer learning clearly improves the generation quality and diversity over the models trained from scratch only with the target dataset.
\figSynthEx

Unfortunately, the effectiveness of transfer learning for GANs highly depends on how similar the source dataset is to the target dataset. According to TransferGAN~\cite{wang2018transfergan}, transfer learning can achieve the best performance when the source shares common characteristics with the target. For example, when LFW~\cite{huang2014LFW} is the target dataset, the best performance is achieved with the source dataset of CelebA~\cite{liu2015celeba} as both are face datasets. For Flower~\cite{nilsback2008flowers} or Kitchens~\cite{yu15lsun}, utilizing CelebA as the source dataset does not significantly improve the performance. Thus, it is required to search the best source dataset for a given target dataset by measuring the similarity between two datasets (e.g., FID score). Because exploring the best source dataset and then acquiring its pretrained model is ad-hoc, the search result does not guarantee the best pair for transfer learning~\cite{wang2018transfergan}. Moreover, none of the existing source datasets can sufficiently fit the target dataset in real-world applications.

Other than the performance issue, we argue that the current pretrained models can be vulnerable to copyright (see the supplementary \appnumCopyright for potential copyright issues of large-scale datasets) and privacy issues~\cite{zou2020privacy}. 
Even for public benchmark datasets, employing them for commercial purposes is not always permitted. 
For examples, ImageNet-1K having 1M images, the copyright issue might not be feasible to handle. When targeting the commercial use of a dataset, the developer should negotiate with the author of each sample. For this reason, one might compose her own dataset via web crawling, but filtering out the copyrighted samples is practically difficult. Besides, unresolved copyright and privacy issues might cause legal issues~\cite{nytimes2020facebook}. 

Recent studies~\cite{hayes2019logan,hilprecht2019monte,chen2020ganleak} also show that the deep generative models are vulnerable to membership inference attacks, implying that privacy issues still remains beyond the copyright issues. An adversary can reconstruct a face even without additional prior information~\cite{zhang2020secret}. That is, we can reveal individual training samples by attacking the trained model. As the network capacity of GANs increases rapidly to improve performance, the risk of memorization also grows quickly. Memorization effects make GANs more vulnerable to membership inference attacks~\cite{carlini2019secret}. Since we consider transfer learning, someone might argue that the membership inference on the source (e.g., pretraining) dataset should not be a critical issue. However, Zou et al.~\cite{zou2020privacy} reported that the membership inference of the source dataset could be conducted even after the transfer learning (see the supplementary \appnumCopyright for empirical evidence).

In this work, we dive into tackling the two undiscovered but critical issues of transfer learning for GANs: 1) the lack of generalization for the pretrained model and 2) the copyright or privacy issue of the pretraining dataset. To this end, we devise a synthetic data generation strategy for acquiring pretrained GANs. Since our pretrained model is newly computed with a synthetic dataset, it is inherently free from copyright and privacy issues. Besides, the learned features of existing pretrained models encode the inductive bias of a training dataset, exhibiting lower transferability~\cite{yosinski2014transferable}. Learned from this lesson, we ensure that our synthetic data should be unbiased to any datasets and free from expert knowledge or specific domain prior.

Towards this goal, we adopt the generic property of the natural images in the frequency spectrum and structure. We develop our data generation strategy, namely \primitivesPS, inspired by the analysis and observations on natural images from previous studies~\cite{oppenheim1981importance,tadmor1993both,mehra2009abstraction}. Our design philosophy is built upon three aspects: 1) considering the power spectrum distribution of the natural images as in \Cref{fig:synthetic_example}(a), 2) reflecting the structural property of the natural images as illustrated in \Cref{fig:synthetic_example}(b), and 3) utilizing the existence of saliency in images (\Cref{fig:synthetic_example}(c) shows the synthetic data generated by applying both 2) and 3).) Finally, we combine all three aspects and develop our final data synthesizer \primitivesPS, as visualized in \Cref{fig:synthetic_example}(d). We pretrain GANs using the synthetic dataset generated by our data synthesizer. Then, the effectiveness of the proposed method is evaluated by repurposing the pretrained model to various low-shot datasets.

Extensive evaluations and analysis confirm that this single pretrained network 1) can be effectively transferred to various low-shot datasets and 2) improve the generation performance and the convergence time. Interestingly, the model pretrained with our dataset outperforms the model pretrained with the natural images when transferred to several datasets. Our empirical study shows that the bias from a specific dataset for pretraining GANs is harmful to the generalization performance of transfer learning. Finally, our analysis of learned filters provides insight into what makes the pretrained model transferable. The code is available at {\footnotesize \url{https://github.com/FriedRonaldo/Primitives-PS}}.

\section{Related work}

\label{sec:related_work}
\subsection{Utilizing synthetic datasets}
The samples and labels of synthetic datasets can be generated automatically and unlimitedly by a pre-defined process. Since generating synthetic data can bypass the cumbersome data crawling and pruning for data collection, previous works have utilized synthetic datasets for training the model and then achieved performance improvement on real datasets~\cite{hou2021visualizing,hoffman2018cycada,shu2018a,roy2021curriculum,saito2018maximum,yang2020fda,tobin2017domainrand}. Domain randomization~\cite{tobin2017domainrand} used various illuminations, color, noise, and texture to reduce the performance gap between the simulated and real samples. By doing so, a model trained with a synthetic dataset helps improve the performance on the real dataset. Fourier domain adaptation~\cite{yang2020fda} proposed swapping the low-frequency components of the synthetic and real samples to reduce the domain gap in the texture. 

Although the previous methods improved the performance of the model on the real dataset, generating such synthetic datasets requires expertise in domain knowledge or a specific software (e.g., GTA-5 game engine~\cite{Richter_2016_GTA}). To handle the issue, Kataoka et al.~\cite{Kataoka_2020_ACCV} utilized the iterated function system to generate fractals and used the fractals as a pretraining dataset for classification. As a concurrent work, Baradad et al.~\cite{baradad2021learning} observe that the unsupervised representation learning~\cite{he2020momentum} trains the model using patches, and these patches are visually similar to the noise patches (from the noise generation model) or the patches drawn from GANs. Based on the observation, they generate synthetic datasets and conduct self-supervised learning for an image classification task. However, none of the existing studies have investigated synthetic data generation for training GANs.  

\subsection{Transfer learning in GANs}
GANs involve a unique architecture and a training strategy; consisting of a discriminator and generator trained via adversarial competition. Therefore, the GAN transfer learning method should be developed by considering the unique characteristics of GANs~\cite{wang2018transfergan,noguchi2019image,wang2020minegan,mo2020freezed,zhao2020leveraging,ojha2021few}. TransferGAN~\cite{wang2018transfergan} trains GANs with a small number of samples by transferring the weights trained on a relatively large dataset. TransferGAN also shows that the performance of the transferred model depends on the relationship between the source and target datasets. Noguchi and  Harada~\cite{noguchi2019image} proposed to update only the statistics of the batch normalization layer for transferring GANs. This strategy prevents GANs from overfitting so that the model can generate diverse images even with a small number of samples. FreezeD~\cite{mo2020freezed} fixes several layers of the discriminator and then finetunes the remaining layers. FreezeD improved the generation performance of transferring from the FFHQ pretrained model to various animals. Despite the improvement in GAN transfer learning, the model still requires a large-scale pretraining dataset. Consequently, they commonly suffer from copyright issues, and their performance is sensitive to the relationship between the source and target dataset. In contrast, our goal is to tackle both issues simultaneously by introducing an effective data synthesizer. 

\subsection{Low-shot learning in GANs}
For high-quality image generation, GANs require a large-scale dataset, and such a requirement can limit the practical use of GANs. To reduce the number of samples for training, several recent studies have introduced data augmentation for training the discriminator\cite{tran2021data,zhao2020diffaugment,Neurips2020Ada}. Then, the generator can produce images with a small number of samples without reflecting an unwanted transformation such as cutout~\cite{devries2017cutout} in the results (i.e., augmentation leakage~\cite{Neurips2020Ada}). Recently, ReMix~\cite{cao2021remix} utilizes interpolation in the style space to reduce the required images to train an image-to-image translation model. In this work, we tackle low-shot generation using GANs via transfer learning; GANs are trained with a small number of samples by transferring a pretrained network into a low-shot dataset.

\section{Towards an effective data synthesizer}
\label{sec:method}
In this work, our primary goal is to develop an unbiased and effective data synthesizer. The synthetic dataset secured by our synthesizer is then used to pretrain GANs, which facilitates low-shot data generation. To accomplish unbiased data generation, we only consider the generic properties of natural images because the inductive bias in a pretraining dataset is harmful to transfer learning of GANs. In the following, we introduce three design philosophies of our data synthesizer inspired by the common characteristics of natural images: 1) learning the power spectrum of natural images, 2) exploiting the shape primitives from natural images, and 3) adopting the existence of saliency in images. 

\subsection{Learning the power spectrum of natural images}
Several previous works reported the magnitude of natural images in the frequency domain ~\cite{field1987relations,burton1987color,tolhurst1992amplitude} roughly obeys $w_m = \frac{1}{|f_x|^a + |f_y|^a}$ where $a$ is a constant, well approximated to one. Inspired by this finding, we generate synthetic images by randomly drawing $a$ from the uniform distribution of $\mathcal{U}(0.5, 3.5)$, as also suggested in \cite{baradad2021learning}. Specifically, random white noise is sampled, and then its magnitude signal after applying the Fast Fourier Transform (FFT) is weighted by $w_m$. By applying the inverse FFT to the weighted signal, we can easily compute the synthetic image. We repeat this for RGB color channels and finally produce synthetic images. Originally, the image with $a=1$ was named a pink noise. We call this method of generating images with $a \sim \mathcal{U}(0.5, 3.5)$ as \pinknoise. Since we only utilize the generic properties of natural images, no inductive bias toward any specific dataset influences \pinknoise. As shown in \Cref{fig:synthetic_example}(a), \pinknoise produces interesting patterns with vertical, horizontal orientation, or color blobs.

\figAbs
\subsection{Shape primitives inspired by natural images}
\epigraph{\textit{``Everything in nature is formed upon the sphere, the cone, and the cylinder. One must learn to paint these simple figures, and then one can do all that he may wish.''}}{Paul C\'ezanne}
Considering the importance of phase in images (e.g., determining the unique appearance of the image~\cite{oppenheim1981importance}), \pinknoise alone is insufficient to represent the rich characteristics of natural images; \pinknoise is random noise on a phase spectrum. To have a meaningful signal even in its phase, we can consider 1) modeling the phase of natural images independently or 2) developing the different generation strategies to model the magnitude and phase simultaneously. Unlike the magnitude spectrum, we seldom find regularity in the phase of images; thus, it is difficult to derive the generic property of the phase spectrum. Besides, separately modeling the phase and magnitude may not produce meaningful images, preserving the proper structures~\cite{tadmor1993both}. For this reason, we focus on finding structural regularity in natural images because it can affect both magnitude and phase. Specifically, we are inspired by the observation that natural images can be represented by the composition of the elementary shapes~\cite{mehra2009abstraction}. The common practice in artistic drawings also utilizes elementary shapes as the basis for representing things (inspired by Paul C\'ezanne).

\Cref{fig:abstraction} demonstrates the abstraction examples of various images using elementary shapes, such as ellipses, lines, and rectangles. We find the potential of abstraction via elementary shapes to encode the structural information of natural images and to remove the bias to a specific dataset. We then devise the data synthesizer to produce images consisting of various elementary shapes. The outputs of this synthesis procedure are akin to those of the dead leaves model~\cite{haralick1987image,lee2001occlusion}. The dead leaves model is an early generative model, which closely mimics natural images by conducting tessellation, where their sizes and positions are determined by sampling from the Poisson process. Unlike the dead leaves model, we do not fill all the regions and use different distributions for sampling because the resultant images are quite sensitive to the hyperparameter of the Poisson process. For position, we use the uniform distribution. To prevent the large shapes in the later stage from completely overwriting those in the early stage, we gradually decrease the maximum shape size over multiple stages; drawing the small objects toward the end. In addition, it is conversely proportional to the number of currently injected shapes. We name this generation strategy \primitives, and \Cref{fig:synthetic_example}(b) visualizes the representative examples. By distributing the shapes in the image space, we observe that \primitives produces images that have a similar magnitude to those of natural images (See \Cref{tab:freq_distance} and the supplementary \appnumFFTImage for the supporting experiments).

\tabFreqDist

\subsection{Combining saliency as prior} 

In addition to the natural images, we investigate the benchmark datasets and find that they commonly have saliency, target objects of interest to determine the class. These salient objects are usually located nearly in the middle of the image. For example, the animal face on the cat and panda dataset can be the saliency. To reflect the nature of benchmark datasets, we insert a large shape after applying \primitives and name it as \primitivesS (\primitives with \texttt{S}aliency).

By utilizing the three design factors, we develop four variants of our data synthesizer. They are 
1) \pinknoise adopting the nature of magnitude spectrum of natural images only as shown in \Cref{fig:synthetic_example}(a), 2) \primitives generating various elementary (monotone) shapes randomly as illustrated in \Cref{fig:synthetic_example}(b), and 3) \primitivesS adding a salient object into \primitives in \Cref{fig:synthetic_example}(c). 

\figPrimitivesVSPrimitivesPS
Finally, we apply a \pinknoise pattern onto the salient object and the background of \primitivesS, which is called (4) \primitivesPS (\primitives with \texttt{P}atterned \texttt{S}aliency) as shown in \Cref{fig:synthetic_example}(d). Since the size of the salient object is considerable, having a salient monotone object may induce an unwanted texture bias. Focusing on the visual effects, inserting the monotone object can be similar to the regional dropout~\cite{singh2017HnS,baek2020psynet} in the weakly-supervised object localization (WSOL) task. When training a network with the regional dropout, previous WSOL methods suggest filling the dropped region with mean statistics or with other regions from the same image to prevent distribution bias. Motivated by the practice in WSOL, we apply \pinknoise to the salient object. 

The effectiveness of the proposed synthetic datasets is evaluated by transferring GANs in \Cref{sec:experiment}. We first pretrain GANs using the randomly generated images via our \primitivesPS, and then finetune the pretrained model on low-shot datasets. While finetuning, all competitors and our pretrained model utilize DiffAug (translation, cutout, and color jittering). For the pretraining results and the details, please refer to the supplementary \appnumPretrainingResult.

\section{Experiments}
\label{sec:experiment}
We first demonstrate the effectiveness of four variants of our data synthesizer. Then, we choose the best strategy among the four variants and use it for pretraining GANs. Our pretrained model is compared with other pretrained models using a natural benchmark dataset in the transfer learning scenario. We also provide an ablation study on the number of particles in each synthetic image and a policy to determine the size of each particle in the supplementary \appnumAblation. 

\noindent \textbf{Datasets.} For the comparison between our synthesizers, we adopt four datasets, including Obama, Grumpy cat, Panda, and Bridge of sighs (Bridge)~\cite{zhao2020diffaugment}. To compare with transfer learning methods, we also use Wuzhen, Temple of heaven (Temple), and Medici fountain (Fountain). Each dataset has 100 images. In addition, we create a dataset, namely Buildings, by merging a subset of four datasets; Bridge of sighs, Wuzhen, Temple of heaven, and Medici fountain. Buildings is used to evaluate the performance under highly diverse conditions. For comprehensive evaluations, we also use CIFAR-10/100 datasets when training with BigGAN.

\tabCompareSynth
\noindent \textbf{Evaluation protocols.} StyleGAN2 architecture~\cite{karras2020StyleGAN2} with DiffAug~\cite{zhao2020diffaugment} is applied when evaluating all models in the low-shot generation task. The baseline is the model trained from scratch with DiffAug. The strong competitors are TransferGAN~\cite{wang2018transfergan} and FreezeD~\cite{mo2020freezed}, where both methods suggest finetuning strategies. To reproduce the competitors, we pretrain StyleGAN2 on FFHQ-- the face dataset and then finetune the pretrained model using TransferGAN with DiffAug and FreezeD with DiffAug, respectively. Since the baseline can outperform the competitors upon the target datasets, we report the baseline performances for comparison. Besides, we stress that all competitors, baseline and \primitivesPS use DiffAug. Specifically, we follow the configuration of DiffAug for \primitivesPS and the baseline (from scratch with DiffAug). Otherwise, we use the configuration of TransferGAN and FreezeD as described in \cite{zhao2020diffaugment} for the best performance. 

We also apply our synthetic dataset to pretrain BigGAN~\cite{brock2018biggan} and repurpose the model to CIFAR-10/100 datasets for evaluating our synthesizer in the conditional generation task. Since \primitivesPS does not have labels, we randomly assigned the labels during pretraining. We developed the pretrained model independently for CIFAR-10 and 100 as they have different architectures due to different numbers of classes. For evaluating the conditional generation task, we compare three models; 1) the model na\"ivly trained from scratch, 2) the model trained with DiffAug only (DiffAug), and 3) our model pretrained with \primitivesPS and then finetuned with DiffAug. We use 10\%, 20\%, and 100\% samples of CIFAR for evaluation and check the effectiveness of our strategy under the data-scarce scenario. As an evaluation metric, we use Fr\'echet inception distance (FID)~\cite{heusel2017FID} and report the FID score of the best model during training as suggested by DiffAug~\cite{zhao2020diffaugment}. We also provide KMMD~\cite{xu2018KMMD} for the better quantitative evaluation, please refer to supplementary \appnumKMMD.

\figDiff
\figLowShotQual
\tabCompareTransfer
\subsection{Effects of different data synthesizers}
We developed four variants of data synthesizer: \pinknoise, \primitives, \primitivesS, and \primitivesPS. We evaluate their effectiveness in the low-shot generation scenario-- pretraining with the synthetic dataset and then finetuning on target datasets with DiffAug. 
\Cref{tab:compare_synth} summarizes the FID scores of four data synthesizers and the baseline under four different low-shot datasets. 

In general, \pinknoise fails to improve the FID score (worse than the baseline), but converges fast (See the supplementary \appnumConvergeSpeedSynth). Unlike \pinknoise, \primitives clearly improves the generation performance in Obama and Panda, large margins from the baseline. However, it is not effective on Grumpy cat and Bridge. Compared to \primitives, \primitivesS further improves the FID scores, demonstrating the effectiveness of saliency prior. 
Finally, \primitivesPS clearly improves the low-shot generation performance on all datasets by about 15\% on average over the baseline. We provide the qualitative evaluation in the supplementary \appnumQualitativeSynth. From these results, we observe that 1) a na\"ive synthesizer (\pinknoise) is even worse than simply using the low-shot dataset, and 2) the combination of our three design factors (\primitivesPS) remarkably improves the baseline, supporting the effectiveness and importance of each factor.

To analyze how closely our data synthesizers mimic the real datasets, we focus on measuring the similarity between our synthetic dataset (source) and the actual low-shot dataset (target). Instead of pixel distance, we compare the average structural similarity (SSIM) between two datasets in the frequency domain. Since the phase periodically varies in $[-\pi, \pi]$, the SSIM of the phase spectrum is not reliable for comparison. Therefore, we only report the SSIM using the magnitude spectrum in \Cref{tab:freq_distance}. We confirm that similar trends are consistently observed in L1 or L2 distance. The value of the SSIM is not an exact indicator for explaining the FID scores. Nevertheless, it helps understand the gains; the low-shot generation performance improves as our data synthesizer models the target dataset more similarly. In \Cref{tab:compare_synth}, \primitivesS and \primitivesPS were ranked top-2, except for Obama. The two strategies in \Cref{tab:freq_distance} also show that their magnitude spectrum is the most similar to target datasets. This interesting trend supports that our design factors are effective choices to mimic the statistics of real images.

We also visualize how our synthetic data gradually fit the target data by showing the generation results at different training stages. For that, \primitives and \primitivesPS are selected to construct the pretrained model, and then they are transferred to Obama. By comparing \primitives and \primitivesPS, we observe the effect of the saliency prior. \Cref{fig:primitives_vs_primitivesPS} shows that the salient shape in \primitivesPS forms the main object as the training evolves. Meanwhile, \primitives includes multiple shapes, meaning all can be candidates for the main object. Consequently, the results often contain multiple faces in the middle of training (e.g., the top-left, the top-right, and the middle in \Cref{fig:primitives_vs_primitivesPS}(a)). On the other hand, \primitivesPS focuses on generating a single face and eventually exhibits improved quality. We further visualize the gradual changes in outputs of \primitivesPS pretrained model in \Cref{fig:differentiation}. For the full animation, please refer to the supplementary material (GIF files).

Considering all, we confirm that \primitivesPS is the best data synthesizer, and thus it is chosen as our final model for comparative evaluations with competitors.

\subsection{Comparisons with the state-of-the-arts}
We pretrain a model using \primitivesPS and compare it with state-of-the-art models pretrained with natural images in a transfer learning task to low-shot datasets.

\Cref{tab:compare_transfer} reports the quantitative results and \Cref{fig:lowshot_qual} shows the qualitative comparison. As expected, TransferGAN~\cite{wang2018transfergan} and FreezeD~\cite{mo2020freezed} show outstanding performance on the Obama dataset because they are pretrained with FFHQ, meaning the source dataset is a superset of the target. Except for the Obama dataset, our pretrained model with \primitivesPS outperforms all competitors. Unless the inductive bias in the source dataset is advantageous to the target (e.g., Obama), FreezeD does not consistently outperform the baseline (from scratch with DiffAug). In fact, the performances of existing methods highly vary upon target datasets. Contrarily, our pretrained model with \primitivesPS consistently outperforms the competitors in each dataset, except Obama. This implies that our pretrained model has strong transferability. Since \primitivesPS does not use any inductive bias for modeling human faces, the performance drawback on Obama can be acceptable.

We emphasize that our achievement in generation quality is impressive and meaningful in two aspects: 1) \primitivesPS uses no real but all synthetic images, which possesses all the attractive nature in application scenarios and 2) our results show the great potential of a single pretrained model for GAN transfer learning.

\tabFilterDiv
\tabCIFAR

\noindent\textbf{Diverse filters matter for transferring GANs.} From the superior performances of our pretrained model, we conjecture that our achievement was possible by the unbiased nature of our dataset; the pretrained model with FFHQ (FreezeD) has an inductive bias as the face dataset. A previous study analyzing the transferability of CNN~\cite{yosinski2014transferable} also pointed out that the performance of the target dataset degrades when the filters are highly specialized to the source dataset. To analyze the transferability empirically, we measure the similarity between the filters of each layer of the pretrained model. We regard that highly diverse (less similar to each other) filters can indicate that the model is less biased towards a particular domain. That means that the highly transferable model tends to have low filter similarity on average. Specifically, given a weight matrix of each layer, its shape is $[O, I, H, W]$, where $O$ filters have $I \times H \times W$ tensors. Then, we measure the cosine similarity among all possible permutations of $O$ filters and report the mean value of the average similarity of all layers in \Cref{tab:filter_diversity}. For all the layers, please refer to the supplementary \appnumSimilarityFilter. 

In summary, \primitivesPS shows the more diverse filter set in 21 out of 26 layers than the FFHQ pretrained model. According to \cite{yosinski2014transferable}, the higher layer (close to the output) tends to specialize in the trained dataset. The same observation holds in our discriminator. The similarity in the last layer of the FFHQ pretrained model is approximately four times higher than \primitivesPS. This explains that the FFHQ pretrained model specialized in human faces, thus transferring well to Obama but not to others.

\figFIDperIT
\figCIFAR
\noindent\textbf{Training convergence speed.} We investigate the convergence speed of transfer learning by examining FID upon training iterations. \Cref{fig:fid_per_it} describes the evolution of the FID scores during the training. To save space, we provide two different datasets; Obama and Bridge. Results for the complete set are in the supplementary \appnumConvergeSpeedComp. For Obama, all pretrained models converge faster than the baseline (from scratch with DiffAug). Meanwhile, only our model converges faster than the baseline for Bridge. Compared to the baseline, the model pretrained with \primitivesPS reaches 95\% of the best baseline performance within the first 30\% of iterations. Interestingly, other pretrained models cannot reach 95\% of the best baseline performance earlier than the baseline. This shows that our model effectively reduces the required iterations for convergence, and the overhead for pretraining can be sufficiently deducted.

\noindent\textbf{Toward a conditional generation task using CIFAR.} We conduct conditional generation via transfer learning on CIFAR-10 and 100 as summarized in \Cref{tab:cifar}. \Cref{fig:cifar} shows the qualitative evaluation result on CIFAR-10 with 10\% of samples; our \primitivesPS produces the general shape and its structural components better than the baseline and DiffAug. Compared to BigGAN trained from scratch, BigGAN trained from scratch with DiffAug significantly improves the FID score, and the gain is pronounced as the number of training samples decreases. However, we observe that DiffAug suffers from augmentation leakage~\cite{Neurips2020Ada} when the samples are scarce (i.e., the generated samples contain the cutout). Our pretrained model with \primitivesPS shows remarkable performances under the data-hungry scenario, better than DiffAug.

However, when the samples are sufficient (100\%), pretraining does not always provide gains over DiffAug. This tendency appears in various downstream tasks. Newell et. al.~\cite{newell2020useful} reported that the self-supervised pretraining for semi-supervised classification is not advantageous when the amount of data-label pairs are sufficient. TransferGAN~\cite{wang2018transfergan} showed that the gain via transfer learning decreases when the amount of samples is sufficient. In the same vein, the advantage of our pretraining with \primitivesPS decreases as the number of samples increases. 

For the extreme low-shot scenario, we also evaluated the model trained with 1\% of the dataset. Only for this evaluation, we compare three models; 1) the model na\"ivly trained from scratch, 2) the model trained with DiffAug only (DiffAug), and 3) our model pretrained with \primitivesPS and then finetuned without DiffAug. The FID score of the baseline, DiffAug, and ours are 112.13, 101.91, and 78.48, respectively. Although DiffAug improved FID, we observe that DiffAug suffers from the augmentation leakage issue. Therefore, the improvement in FID and its generation results are not meaningful. In contrast, our pretrained model can significantly improve the generation performance without any issue. For more details and results for CIFAR, please refer to the supplementary \appnumCIFAR.

\section{Discussion and conclusion}
\label{sec:discussion}

\noindent\textbf{Societal impact.} Since we propose the synthetic dataset for pretraining, the proposed method can consume more power at the pretraining stage (generating the synthetic data and then pretraining the model). However, it converges much faster for finetuning on target datasets, and the same model can be repeatedly used for all targets. In this regard, our method is eventually the more efficient choice in terms of power consumption.
In the point of the ethical view, especially considering the bias issues (e.g., racial or gender bias) in the current benchmark datasets, using our method is much more safe, fair, economical, and practical. Besides, pretraining with our synthetic dataset guarantees the robustness of membership inference attack towards the source dataset because reconstructing our data is meaningless. Since our method is copyright-free, it helps small commercial groups to develop their machine-learning model. 

\noindent\textbf{Limitation.} Our \primitivesPS is devised based on the observations from natural images. Hence, it is possible that more effective observations can further improve the data generation quality. In future work, we plan to develop a metric to quantify the transferability of the model and then derive the data generation process by optimizing the transferability. Formulating such a metric will be challenging but constructive for predicting the behavior of the pretrained model and practically useful in various applications.

\noindent\textbf{Conclusion.} Existing studies for GAN transfer learning utilize a model trained with natural images and thereby suffer from 1) biased pretrained model that can be harmful to the resultant performance and 2) copyright or privacy issues with both the model and dataset. To overcome these limitations, we introduce a new image synthesizer, namely \primitivesPS, inspired by the three generic properties of natural images: 1) following the power spectrum of natural images, 2) abstracting the image via the composition of primitive shapes (e.g., line, circle, and rectangle), and 3) having saliency in the image. Experimental comparisons and analysis show that our strategy effectively improves both the generation quality and the convergence speed. We further investigate the diversity of learned filters and report that they are meaningful evidence for discovering the transferability of the pretrained model. 

{\footnotesize \noindent\textbf{Acknowledgements.}
We thank Jongwuk Lee and CVML members for the valuable feedback. This research was supported by the NRF Korea funded by the MSIT (2022R1A2C3011154, 2020R1A4A1016619), the IITP grant funded by the MSIT (2020-0-01361, YONSEI UNIVERSITY), and the Korea Medical Device Development Fund grant (202011D06).}




{\small
\bibliographystyle{ieee_fullname}
\bibliography{egbib}

\begin{thebibliography}{10}\itemsep=-1pt

\bibitem{baek2021tunit}
Kyungjune Baek, Yunjey Choi, Youngjung Uh, Jaejun Yoo, and Hyunjung Shim.
\newblock Rethinking the truly unsupervised image-to-image translation.
\newblock In {\em Proceedings of the IEEE/CVF International Conference on
  Computer Vision}, pages 14154--14163, 2021.

\bibitem{baek2020psynet}
Kyungjune Baek, Minhyun Lee, and Hyunjung Shim.
\newblock Psynet: Self-supervised approach to object localization using point
  symmetric transformation.
\newblock In {\em Proceedings of the AAAI Conference on Artificial
  Intelligence}, volume~34, pages 10451--10459, 2020.

\bibitem{baradad2021learning}
Manel Baradad, Jonas Wulff, Tongzhou Wang, Phillip Isola, and Antonio Torralba.
\newblock Learning to see by looking at noise.
\newblock In {\em Advances in Neural Information Processing Systems}, 2021.

\bibitem{brock2018biggan}
Andrew Brock, Jeff Donahue, and Karen Simonyan.
\newblock Large scale {GAN} training for high fidelity natural image synthesis.
\newblock In {\em International Conference on Learning Representations}, 2019.

\bibitem{burton1987color}
Geoffrey~J Burton and Ian~R Moorhead.
\newblock Color and spatial structure in natural scenes.
\newblock {\em Applied optics}, 26(1):157--170, 1987.

\bibitem{cao2021remix}
Jie Cao, Luanxuan Hou, Ming-Hsuan Yang, Ran He, and Zhenan Sun.
\newblock Remix: Towards image-to-image translation with limited data.
\newblock In {\em Proceedings of the IEEE/CVF Conference on Computer Vision and
  Pattern Recognition}, pages 15018--15027, 2021.

\bibitem{carlini2019secret}
Nicholas Carlini, Chang Liu, {\'U}lfar Erlingsson, Jernej Kos, and Dawn Song.
\newblock The secret sharer: Evaluating and testing unintended memorization in
  neural networks.
\newblock In {\em 28th $\{$USENIX$\}$ Security Symposium ($\{$USENIX$\}$
  Security 19)}, pages 267--284, 2019.

\bibitem{chen2020ganleak}
Dingfan Chen, Ning Yu, Yang Zhang, and Mario Fritz.
\newblock Gan-leaks: A taxonomy of membership inference attacks against
  generative models.
\newblock In {\em Proceedings of the 2020 ACM SIGSAC conference on computer and
  communications security}, pages 343--362, 2020.

\bibitem{choe2017face}
Junsuk Choe, Song Park, Kyungmin Kim, Joo Hyun~Park, Dongseob Kim, and Hyunjung
  Shim.
\newblock Face generation for low-shot learning using generative adversarial
  networks.
\newblock In {\em Proceedings of the IEEE International Conference on Computer
  Vision Workshops}, pages 1940--1948, 2017.

\bibitem{choi2020starganv2}
Yunjey Choi, Youngjung Uh, Jaejun Yoo, and Jung-Woo Ha.
\newblock Stargan v2: Diverse image synthesis for multiple domains.
\newblock In {\em Proceedings of the IEEE/CVF Conference on Computer Vision and
  Pattern Recognition}, pages 8188--8197, 2020.

\bibitem{devries2017cutout}
Terrance DeVries and Graham~W Taylor.
\newblock Improved regularization of convolutional neural networks with cutout.
\newblock {\em arXiv preprint arXiv:1708.04552}, 2017.

\bibitem{field1987relations}
David~J Field.
\newblock Relations between the statistics of natural images and the response
  properties of cortical cells.
\newblock {\em Josa a}, 4(12):2379--2394, 1987.

\bibitem{2014GANs}
Ian Goodfellow, Jean Pouget-Abadie, Mehdi Mirza, Bing Xu, David Warde-Farley,
  Sherjil Ozair, Aaron Courville, and Yoshua Bengio.
\newblock Generative adversarial nets.
\newblock In {\em Advances in neural information processing systems}, pages
  2672--2680, 2014.

\bibitem{haralick1987image}
Robert~M Haralick, Stanley~R Sternberg, and Xinhua Zhuang.
\newblock Image analysis using mathematical morphology.
\newblock {\em IEEE transactions on pattern analysis and machine intelligence},
  (4):532--550, 1987.

\bibitem{hayes2019logan}
Jamie Hayes, Luca Melis, George Danezis, and Emiliano De~Cristofaro.
\newblock Logan: Membership inference attacks against generative models.
\newblock In {\em Proceedings on Privacy Enhancing Technologies (PoPETs)},
  volume 2019, pages 133--152. De Gruyter, 2019.

\bibitem{he2020momentum}
Kaiming He, Haoqi Fan, Yuxin Wu, Saining Xie, and Ross Girshick.
\newblock Momentum contrast for unsupervised visual representation learning.
\newblock In {\em Proceedings of the IEEE/CVF Conference on Computer Vision and
  Pattern Recognition}, pages 9729--9738, 2020.

\bibitem{heusel2017FID}
Martin Heusel, Hubert Ramsauer, Thomas Unterthiner, Bernhard Nessler, and Sepp
  Hochreiter.
\newblock Gans trained by a two time-scale update rule converge to a local nash
  equilibrium.
\newblock {\em Advances in neural information processing systems}, 30, 2017.

\bibitem{hilprecht2019monte}
Benjamin Hilprecht, Martin H{\"a}rterich, and Daniel Bernau.
\newblock Monte carlo and reconstruction membership inference attacks against
  generative models.
\newblock {\em Proc. Priv. Enhancing Technol.}, 2019(4):232--249, 2019.

\bibitem{hoffman2018cycada}
Judy Hoffman, Eric Tzeng, Taesung Park, Jun-Yan Zhu, Phillip Isola, Kate
  Saenko, Alexei Efros, and Trevor Darrell.
\newblock Cycada: Cycle-consistent adversarial domain adaptation.
\newblock In {\em International conference on machine learning}, pages
  1989--1998. PMLR, 2018.

\bibitem{hou2021visualizing}
Yunzhong Hou and Liang Zheng.
\newblock Visualizing adapted knowledge in domain transfer.
\newblock In {\em Proceedings of the IEEE/CVF Conference on Computer Vision and
  Pattern Recognition}, pages 13824--13833, 2021.

\bibitem{huang2014LFW}
Gary~B Huang and Erik Learned-Miller.
\newblock Labeled faces in the wild: Updates and new reporting procedures.
\newblock {\em Dept. Comput. Sci., Univ. Massachusetts Amherst, Amherst, MA,
  USA, Tech. Rep}, 14(003), 2014.

\bibitem{Neurips2020Ada}
Tero Karras, Miika Aittala, Janne Hellsten, Samuli Laine, Jaakko Lehtinen, and
  Timo Aila.
\newblock Training generative adversarial networks with limited data.
\newblock In H. Larochelle, M. Ranzato, R. Hadsell, M.~F. Balcan, and H. Lin,
  editors, {\em Advances in Neural Information Processing Systems}, volume~33,
  pages 12104--12114. Curran Associates, Inc., 2020.

\bibitem{karras2020StyleGAN2}
Tero Karras, Samuli Laine, Miika Aittala, Janne Hellsten, Jaakko Lehtinen, and
  Timo Aila.
\newblock Analyzing and improving the image quality of stylegan.
\newblock In {\em Proceedings of the IEEE/CVF Conference on Computer Vision and
  Pattern Recognition}, pages 8110--8119, 2020.

\bibitem{Kataoka_2020_ACCV}
Hirokatsu Kataoka, Kazushige Okayasu, Asato Matsumoto, Eisuke Yamagata, Ryosuke
  Yamada, Nakamasa Inoue, Akio Nakamura, and Yutaka Satoh.
\newblock Pre-training without natural images.
\newblock In {\em Proceedings of the Asian Conference on Computer Vision
  (ACCV)}, November 2020.

\bibitem{ledig2017SRGAN}
Christian Ledig, Lucas Theis, Ferenc Husz{\'a}r, Jose Caballero, Andrew
  Cunningham, Alejandro Acosta, Andrew Aitken, Alykhan Tejani, Johannes Totz,
  Zehan Wang, et~al.
\newblock Photo-realistic single image super-resolution using a generative
  adversarial network.
\newblock In {\em Proceedings of the IEEE conference on computer vision and
  pattern recognition}, pages 4681--4690, 2017.

\bibitem{lee2001occlusion}
Ann~B Lee, David Mumford, and Jinggang Huang.
\newblock Occlusion models for natural images: A statistical study of a
  scale-invariant dead leaves model.
\newblock {\em International Journal of Computer Vision}, 41(1):35--59, 2001.

\bibitem{li2018layoutgan}
Jianan Li, Tingfa Xu, Jianming Zhang, Aaron Hertzmann, and Jimei Yang.
\newblock Layout{GAN}: Generating graphic layouts with wireframe discriminator.
\newblock In {\em International Conference on Learning Representations}, 2019.

\bibitem{liu2015celeba}
Ziwei Liu, Ping Luo, Xiaogang Wang, and Xiaoou Tang.
\newblock Deep learning face attributes in the wild.
\newblock In {\em Proceedings of International Conference on Computer Vision
  (ICCV)}, December 2015.

\bibitem{mehra2009abstraction}
Ravish Mehra, Qingnan Zhou, Jeremy Long, Alla Sheffer, Amy Gooch, and Niloy~J
  Mitra.
\newblock Abstraction of man-made shapes.
\newblock In {\em ACM SIGGRAPH Asia 2009 papers}, pages 1--10. 2009.

\bibitem{mescheder2018R1reg}
Lars Mescheder, Andreas Geiger, and Sebastian Nowozin.
\newblock Which training methods for gans do actually converge?
\newblock In {\em International conference on machine learning}, pages
  3481--3490. PMLR, 2018.

\bibitem{mo2020freezed}
Sangwoo Mo, Minsu Cho, and Jinwoo Shin.
\newblock Freeze the discriminator: a simple baseline for fine-tuning gans.
\newblock In {\em CVPR AI for Content Creation Workshop}, 2020.

\bibitem{newell2020useful}
Alejandro Newell and Jia Deng.
\newblock How useful is self-supervised pretraining for visual tasks?
\newblock In {\em Proceedings of the IEEE/CVF Conference on Computer Vision and
  Pattern Recognition}, pages 7345--7354, 2020.

\bibitem{nilsback2008flowers}
Maria-Elena Nilsback and Andrew Zisserman.
\newblock Automated flower classification over a large number of classes.
\newblock In {\em 2008 Sixth Indian Conference on Computer Vision, Graphics \&
  Image Processing}, pages 722--729. IEEE, 2008.

\bibitem{noguchi2019image}
Atsuhiro Noguchi and Tatsuya Harada.
\newblock Image generation from small datasets via batch statistics adaptation.
\newblock In {\em Proceedings of the IEEE/CVF International Conference on
  Computer Vision}, pages 2750--2758, 2019.

\bibitem{ojha2021few}
Utkarsh Ojha, Yijun Li, Jingwan Lu, Alexei~A Efros, Yong~Jae Lee, Eli
  Shechtman, and Richard Zhang.
\newblock Few-shot image generation via cross-domain correspondence.
\newblock In {\em Proceedings of the IEEE/CVF Conference on Computer Vision and
  Pattern Recognition}, pages 10743--10752, 2021.

\bibitem{oppenheim1981importance}
Alan~V Oppenheim and Jae~S Lim.
\newblock The importance of phase in signals.
\newblock {\em Proceedings of the IEEE}, 69(5):529--541, 1981.

\bibitem{park2019spade}
Taesung Park, Ming-Yu Liu, Ting-Chun Wang, and Jun-Yan Zhu.
\newblock Semantic image synthesis with spatially-adaptive normalization.
\newblock In {\em Proceedings of the IEEE/CVF Conference on Computer Vision and
  Pattern Recognition}, pages 2337--2346, 2019.

\bibitem{Richter_2016_GTA}
Stephan~R. Richter, Vibhav Vineet, Stefan Roth, and Vladlen Koltun.
\newblock Playing for data: {G}round truth from computer games.
\newblock In Bastian Leibe, Jiri Matas, Nicu Sebe, and Max Welling, editors,
  {\em European Conference on Computer Vision (ECCV)}, volume 9906 of {\em
  LNCS}, pages 102--118. Springer International Publishing, 2016.

\bibitem{roy2021curriculum}
Subhankar Roy, Evgeny Krivosheev, Zhun Zhong, Nicu Sebe, and Elisa Ricci.
\newblock Curriculum graph co-teaching for multi-target domain adaptation.
\newblock In {\em Proceedings of the IEEE/CVF Conference on Computer Vision and
  Pattern Recognition}, pages 5351--5360, 2021.

\bibitem{saito2018maximum}
Kuniaki Saito, Kohei Watanabe, Yoshitaka Ushiku, and Tatsuya Harada.
\newblock Maximum classifier discrepancy for unsupervised domain adaptation.
\newblock In {\em Proceedings of the IEEE conference on computer vision and
  pattern recognition}, pages 3723--3732, 2018.

\bibitem{shu2018a}
Rui Shu, Hung Bui, Hirokazu Narui, and Stefano Ermon.
\newblock A {DIRT}-t approach to unsupervised domain adaptation.
\newblock In {\em International Conference on Learning Representations}, 2018.

\bibitem{nytimes2020facebook}
Natasha Singer and Mike Isaac.
\newblock Facebook to pay \$550 million to settle facial recognition suit.
\newblock {\em The New York Times}, 2019.

\bibitem{singh2017HnS}
Krishna~Kumar Singh and Yong~Jae Lee.
\newblock Hide-and-seek: Forcing a network to be meticulous for
  weakly-supervised object and action localization.
\newblock In {\em International Conference on Computer Vision (ICCV)}, 2017.

\bibitem{tadmor1993both}
Y Tadmor and DJ Tolhurst.
\newblock Both the phase and the amplitude spectrum may determine the
  appearance of natural images.
\newblock {\em Vision research}, 33(1):141--145, 1993.

\bibitem{tobin2017domainrand}
Josh Tobin, Rachel Fong, Alex Ray, Jonas Schneider, Wojciech Zaremba, and
  Pieter Abbeel.
\newblock Domain randomization for transferring deep neural networks from
  simulation to the real world.
\newblock In {\em 2017 IEEE/RSJ international conference on intelligent robots
  and systems (IROS)}, pages 23--30. IEEE, 2017.

\bibitem{tolhurst1992amplitude}
DJ Tolhurst, Y\_ Tadmor, and Tang Chao.
\newblock Amplitude spectra of natural images.
\newblock {\em Ophthalmic and Physiological Optics}, 12(2):229--232, 1992.

\bibitem{tran2021data}
Ngoc-Trung Tran, Viet-Hung Tran, Ngoc-Bao Nguyen, Trung-Kien Nguyen, and
  Ngai-Man Cheung.
\newblock On data augmentation for gan training.
\newblock {\em IEEE Transactions on Image Processing}, 30:1882--1897, 2021.

\bibitem{wang2020minegan}
Yaxing Wang, Abel Gonzalez-Garcia, David Berga, Luis Herranz, Fahad~Shahbaz
  Khan, and Joost van~de Weijer.
\newblock Minegan: effective knowledge transfer from gans to target domains
  with few images.
\newblock In {\em Proceedings of the IEEE/CVF Conference on Computer Vision and
  Pattern Recognition}, pages 9332--9341, 2020.

\bibitem{wang2018transfergan}
Yaxing Wang, Chenshen Wu, Luis Herranz, Joost van~de Weijer, Abel
  Gonzalez-Garcia, and Bogdan Raducanu.
\newblock Transferring gans: generating images from limited data.
\newblock In {\em Proceedings of the European Conference on Computer Vision
  (ECCV)}, pages 218--234, 2018.

\bibitem{xu2018KMMD}
Qiantong Xu, Gao Huang, Yang Yuan, Chuan Guo, Yu Sun, Felix Wu, and Kilian
  Weinberger.
\newblock An empirical study on evaluation metrics of generative adversarial
  networks.
\newblock {\em arXiv preprint arXiv:1806.07755}, 2018.

\bibitem{yang2020fda}
Yanchao Yang and Stefano Soatto.
\newblock Fda: Fourier domain adaptation for semantic segmentation.
\newblock In {\em Proceedings of the IEEE/CVF Conference on Computer Vision and
  Pattern Recognition}, pages 4085--4095, 2020.

\bibitem{yosinski2014transferable}
Jason Yosinski, Jeff Clune, Yoshua Bengio, and Hod Lipson.
\newblock How transferable are features in deep neural networks?
\newblock {\em Advances in Neural Information Processing Systems},
  27:3320--3328, 2014.

\bibitem{yu15lsun}
Fisher Yu, Yinda Zhang, Shuran Song, Ari Seff, and Jianxiong Xiao.
\newblock Lsun: Construction of a large-scale image dataset using deep learning
  with humans in the loop.
\newblock {\em arXiv preprint arXiv:1506.03365}, 2015.

\bibitem{zhang2019ranksrgan}
Wenlong Zhang, Yihao Liu, Chao Dong, and Yu Qiao.
\newblock Ranksrgan: Generative adversarial networks with ranker for image
  super-resolution.
\newblock In {\em Proceedings of the IEEE/CVF International Conference on
  Computer Vision}, pages 3096--3105, 2019.

\bibitem{zhang2020secret}
Yuheng Zhang, Ruoxi Jia, Hengzhi Pei, Wenxiao Wang, Bo Li, and Dawn Song.
\newblock The secret revealer: Generative model-inversion attacks against deep
  neural networks.
\newblock In {\em Proceedings of the IEEE/CVF Conference on Computer Vision and
  Pattern Recognition}, pages 253--261, 2020.

\bibitem{zhao2020leveraging}
Miaoyun Zhao, Yulai Cong, and Lawrence Carin.
\newblock On leveraging pretrained gans for generation with limited data.
\newblock In {\em International Conference on Machine Learning}, pages
  11340--11351. PMLR, 2020.

\bibitem{zhao2020diffaugment}
Shengyu Zhao, Zhijian Liu, Ji Lin, Jun-Yan Zhu, and Song Han.
\newblock Differentiable augmentation for data-efficient gan training.
\newblock In {\em Conference on Neural Information Processing Systems
  (NeurIPS)}, 2020.

\bibitem{zou2020privacy}
Yang Zou, Zhikun Zhang, Michael Backes, and Yang Zhang.
\newblock Privacy analysis of deep learning in the wild: Membership inference
  attacks against transfer learning.
\newblock {\em arXiv preprint arXiv:2009.04872}, 2020.

\end{thebibliography}
}

\clearpage
\setcounter{section}{0}
\setcounter{figure}{0}
\setcounter{table}{0}
\newcommand{\supplesection}{\section}
\newcommand{\supplesubsection}{\subsection}


\supplesection{Ablation Study}
When developing \primitivesPS, we introduce two hyperparameters; 1) the total number of shapes and 2) the policy to determine the size of each component. For determining the size, we consider three policies; \textbf{Fix}, \textbf{Rand} and \textbf{Decay}. \textbf{Fix} indicates that all particles have the same size. To examine the effect of various scale, we set this size as $H \cdot [\sfrac{1}{10}, \sfrac{1}{5}, \sfrac{1}{2}]$, where $H$ is the image resolution. \textbf{Rand} randomly samples the size from the uniform distribution. Both policies can induce the occlusion of the previously injected shapes by the later shape. \textbf{Decay} can bypass the occlusion issue effectively. \textbf{Decay} arbitrarily samples the size from the uniform distribution, where the maximum size is limited to ($H \cdot \sfrac{1}{5} \cdot \sfrac{(N - n)}{N}$), and $N$ and $n$ are the total number of shapes and the number of previously injected particles. In this way, we can ensure that the shapes inserted in the early stage are still visible in the final data. The upper-side of \Cref{tab:ablation_merge} summarizes the FID score for each policy on four datasets. The differences in FID among \textbf{Fix} policies are trivial in that their ratios are not highly correlated with their ranks. Also, we observe that the shapes at the final stage overwrite the previous shapes. Then, the overall appearance with \textbf{Fix} are similar to \pinknoise with a salient object. We investigate the synthesizer that combines \pinknoise with \texttt{PS} by injecting a saliency and then applying \pinknoise on it. Interestingly, we observe that it shows the similar FID scores to \textbf{Fix}. For \textbf{Rand}, it improves the FID score on Obama and bridge, however, the overall performance is much worse than \textbf{Decay}. Therefore, we choose a \textbf{Decay} policy as default for choosing the size. 

\tabAblationMerge
Besides, the total number of shapes is important because it affects the transferability and the time complexity of the synthesizer. The lower-side of \Cref{tab:ablation_merge} demonstrates the performance trends upon the total number of shapes. A zero particle case implies that only one background and one salient object, thus equivalent to \pinknoise + \texttt{PS}. As the number of shapes ($N$) grows upon roughly 100, the performance tends to improve. However, over $N=100$, we do not observe the consistent gain. From the ablation study, we decide $N=100$ in each image to enjoy the reasonable performance gain and to reduce the time complexity.

\figFIDperITSynthAppendix
\supplesection{Convergence speed of synthetic datasets}
\Cref{fig:fid_per_it_app} shows the evolution of the FID scores during the training of the models pretrained with synthetic datasets. Even if \pinknoise does not improve the generation performance, it can boost the convergence speed. In general, the pretrained models reach 95\% of the best FID score of the from scratch model with DiffAug within first 30\% iterations. The faster convergence speed informs us the positive potential of the pretraining.
\newpage
\supplesection{Qualitative comparison among our data synthesizers}
In addition to the quantitative comparison of our data synthesizers, we also qualitatively compare our four variants of the data synthesizer used for quantitative evaluation. From the first to the last row, Bridge of sighs, Obama, Grumpy cat, and Panda. \pinknoise generates the images with unstructured samples (e.g. Obama and Grumpy cat) and the outputs of \primitives on Panda have lower fidelity (e.g. the last three samples). Compared to \pinknoise and \primitives, \primitivesS and \primitivesPS provide plausible samples. Between the last two synthetic datasets, \primitivesS sometimes drops the important factor, for example, the eyes of the cat (6-th column). While \primitivesPS generates more diverse and plausible samples than the other synthetic datasets. 

\clearpage
\figAppSynthQualPNandPrimitives
\clearpage
\figAppSynthQualSandPS
\clearpage
\clearpage

\supplesection{Convergence speed of transfer learning methods}
\Cref{fig:fid_per_it_comp_app} shows the evolution of the FID scores during the training of the transfer learning methods. The model pretrained with our synthetic dataset exhibits comparable or faster convergence than the competitors that are pretrained on FFHQ. Herein, we observe the convergence speed in terms of the number of iterations to reach 95\% of the best FID score of the baseline (from scratch model with DiffAug).
\figFIDperITCompAppendix
\clearpage

\supplesection{Qualitative comparisons with competing transfer learning methods}
In addition to the quantitative comparison, we also provide the qualitative comparisons on eight datasets that are used for quantitative evaluation in the main text. From the first to the last row, Buildings, Bridge of sighs, Obama, Medici fountain, Grumpy cat, Temple of heaven, Panda, and Wuzhen. 
\figAppQualDiffAug
\newpage 
In terms of fidelity of the generated images, our \primitivesPS outperforms the competitors. Especially, Grumpy cat images generated by the competitors often do not contain eyes or have only part of the face.
\clearpage
\figAppQualTransferGAN
\clearpage
\figAppQualFreezeD
\clearpage
\figAppQualOurs
\clearpage

\clearpage

\tabFilterDivApp
\supplesection{Similarity between filters in all layers}
We calculated the cosine similarity in each layer to measure the diversity of learned filters of pretrained models. FFHQ pretrained model exhibits lower diversity in filters. The average similarity at the last layer of FFHQ pretrained model is approximately four times higher than \primitivesPS. The similar tendency is shown in the first layer of each network -- the consine similarity of FFHQ pretrained model is about two times higher than \primitivesPS.
\newpage

\tabMembership
\supplesection{Copyright issue and vulnerability of pretrained model}
When we directly finetune a pretrained model for commercial use, the trained weights of the model might be defined as software and have the CC BY-ND (creative commons license without modification) license. In this case, we can not utilize the model with post-training or should pay the license fee for the model as software. If we want to use the images for non-commercial purposes, we should acquire the credit of each image from the original author. For ImageNet-1K having 1M images, the copyright issue might not be feasible to handle. When targeting the commercial use of a dataset, the developer should negotiate with the author of each sample. Since this process requires much time and cost to complete, it is likely to be an obstacle to the practical usage of the deep learning system.

Even if we solve the copyright issue via negotiation, the leakage of the training data is another problem. Following the recent work~\cite{zou2020privacy}, the source dataset for pretraining a model can be exposed by the membership inference attack even after the transfer learning. \Cref{tab:membership} shows the empirical evidence. The target models are first pretrained on Caltech101 and transferred to three datasets. The higher AUC, the higher accuracy of the membership inference on the source dataset. Although the accuracy is lower than the attack on the target dataset, it warns us to consider the membership inference attack towards the source dataset seriously.

\clearpage

\figAppCifarLeak
\supplesection{Experimental results on CIFAR}
\supplesubsection{Data augmentation leakage}
The previous work~\cite{Neurips2020Ada} reported the ill-behavior of the data augmentation in GANs; augmentation leakage. When the leakage incurs, the unwanted data transformation is reflected in the generated results. For example, the generated images contain cutout augmentation so that some of the fakes have unwanted empty box. When we train BigGAN on CIFAR with 10\% of samples using DiffAug only, we observe that augmentation leakage. Although the leakage is found, the FID score decreases; FID scores can not reflect the problem of leakage. To penalize this unwanted result, we qualitatively exclude the model with leakage when we find the best model. \Cref{fig:app_cifar_leak} shows the generated images by the model trained with DiffAug (FID: 22.54). Many of the outputs have the unwanted gray box that is the result of leakage of the cutout operation, and this is why we exclude the corresponding FID score in Table 5 of the main text.

On the contrary, the model pretrained with \primitivesPS does not suffer from the leakage even if we use DiffAug (\Cref{fig:app_cifar_ours}). It shows that our pretraining dataset is also effective to prevent augmentation leakage and improves the final generation quality.

\newpage
\figAppCifarOurs

\clearpage
\supplesection{Pretraining results and details}
In this section, we provide the outputs of the generator pretrained with \primitivesPS. For pretraining, we train the model during 800K images with batch size = 16, therefore, the total number of iterations is 50K. For finetuing all the models, we train the model during 400K images. The generated (fake) synthetic images are similar to the real synthetic samples as shown in Figure 1 of the main text.

\figAppPrimitivesPSFakes

\newpage

\supplesection{Frequency domain analysis}
We visualize the average magnitude spectrum of all the samples in Bridge of sighs and compare with the average magnitude spectrum of 1000 images generated by \pinknoise and 1000 images generated by \primitives. The figure below demonstrates their magnitude spectrum. We observe that \primitives produces images that have a similar magnitude spectrum to those of natural images.

\figFFT

\newpage
\tabKMMDApp
\supplesection{Kernel Maximum Mean Discrepancy (KMMD)}
Quantitative evaluation with various metrics is helpful to compare the models and understand the aspect. To this end, we also provide KMMD as suggested by Reviewer 1 in the rebuttal. We report FID only in the main text because of the following reason. In Figure 4(a) of \cite{noguchi2019image}, KMMD considers ``\textit{scale\&shift}'' as the best model although ``\textit{Ours}'' provides more plausible results; ``\textit{scale\&shift}'' even failed to produce eye, nose, and mouth. Contrarily, FID ranked ``\textit{Ours}'' as the best, correctly reflecting the perceptual fidelity.
Table~\ref{tab:KMMD} shows the KMMD score of each model. Although the rankings with KMMD are slightly different from those with FID, our method similarly performs or outperforms the baselines. Overall, we conclude that \primitivesPS is still effective for pretraining GANs. 

\newpage
\tabImageNetApp
\supplesection{Scale-up to higher resolution and comparison with ImageNet}
To check the effectiveness of \primitivesPS in the higher resolution, we pretrain StyleGAN2 with \primitivesPS on 512$\times$512, and then transfer to the low-shot datasets. Moreover, we use the ImageNet pretrained model for all competitors to investigate the effect of a diverse and large-scale training dataset. The pretrained file is from the \href{https://twitter.com/theshawwn/status/1244081581347598341}{link}. We note that this model is pretrained on the 512$\times$512 ImageNet until 1.3M steps. Since the ImageNet dataset can be considered as a super-set of eight test categories, the best performance using the ImageNet pretrained model is often better than \primitivesPS pretrained model. However, when the category of test set no longer overlaps with the ImageNet, we argue that only \primitivesPS can provide consistent and meaningful performances, \emph{e.g., medical images for diagnoses, microscopic images for gene analysis or space imaging for navigation}. Besides, the pretrained model with the 1M ImageNet dataset is vulnerable to the private and copyright issue. A number of images contain a person and the copyright of each image might not be free to all the users. For these practical issues related to legality, the proposed \primitivesPS provides huge benefits for pretraining of GANs.

\end{document}